\documentclass[10pt,journal,compsoc]{IEEEtran}
\ifCLASSOPTIONcompsoc
  \usepackage[nocompress]{cite}
\else
  \usepackage{cite}
\fi
\ifCLASSINFOpdf
  \usepackage[pdftex]{graphicx}
\else
\fi
\usepackage{amsmath}
\usepackage{amsfonts}
\usepackage{url}

\usepackage{multirow}
\usepackage{arydshln}
\usepackage{booktabs}
\usepackage{threeparttable}
\usepackage{tikz}
\newcommand*{\circled}[1]{\lower.7ex\hbox{\tikz\draw (0pt, 0pt)
    circle (.5em) node {\makebox[1em][c]{\small #1}};}}

\begin{document}
%
\title{A Multi-task Multi-stage Transitional Training Framework for Neural Chat Translation}
%
%
%
%

\author{Chulun~Zhou$^\ast$\thanks{$\ast$ \ \ \ C. Zhou and Y. Liang equally contribute to this paper.}, Yunlong Liang$^\ast$, Fandong Meng, Jie Zhou, Jinan Xu, \\ Hongji Wang, Min Zhang and~Jinsong Su$^\dag$\thanks{$\dag$ \ \ \ Jinsong Su is the corresponding author.}
\IEEEcompsocitemizethanks{
\IEEEcompsocthanksitem C. Zhou and H. Wang are with School of Informatics, Xiamen University, Xiamen 361005, China.\protect\\
E-mail: clzhou@stu.xmu.edu.cn, whj@xmu.edu.cn
\IEEEcompsocthanksitem Y. Liang and J. Xu are with Beijing Jiaotong University, Beijing 100044, China.\protect\\
E-mail: yunlongliang@bjtu.edu.cn, jaxu@bjtu.edu.cn
\IEEEcompsocthanksitem F. Meng and J. Zhou are with Pattern Recognition Center, WeChat AI, Tencent Inc, China.\protect\\
E-mail: fandongmeng@tencent.com, withtomzhou@tencent.com
\IEEEcompsocthanksitem M. Zhang is with Soochow University 215031, Suzhou, China.\protect\\
E-mail: minzhang@suda.edu.cn.
\IEEEcompsocthanksitem J. Su is with School of Informatics and Institute of Artificial Intelligence, Xiamen University 361005, Xiamen, China. Meanwhile, he is with Laboratory of Digital Protection and Intelligent Processing of Intangible Cultural Heritage of Fujian and Taiwan (Xiamen University), Ministry of Culture and Tourism, China. He is also with Pengcheng Laboratory, China. \protect\\
E-mail: jssu@xmu.edu.cn
}
\thanks{Manuscript received July 25, 2021; revised May 6, 2022; accepted Dec 18, 2022.}}

%
%

\markboth{Journal of \LaTeX\ Class Files,~Vol.~14, No.~8, August~2015}%
{Shell \MakeLowercase{\textit{et al.}}: Bare Demo of IEEEtran.cls for IEEE Journals}
%



\IEEEtitleabstractindextext{%
\begin{abstract}
Neural chat translation (NCT) aims to translate a cross-lingual chat between speakers of different languages. Existing context-aware NMT models cannot achieve satisfactory performances due to the following inherent problems: 1) limited resources of annotated bilingual dialogues; 2) the neglect of modelling conversational properties; 3) training discrepancy between different stages. To address these issues, in this paper, we propose a \textbf{m}ulti-task \textbf{m}ulti-stage \textbf{t}ransitional (MMT) training framework, where an NCT model is trained using the bilingual chat translation dataset and additional monolingual dialogues. We elaborately design two auxiliary tasks, namely utterance discrimination and speaker discrimination, to introduce the modelling of dialogue coherence and speaker characteristic into the NCT model. The training process consists of three stages: 1) sentence-level pre-training on large-scale parallel corpus; 2) intermediate training with auxiliary tasks using additional monolingual dialogues; 3) context-aware fine-tuning with gradual transition. Particularly, the second stage serves as an intermediate phase that alleviates the training discrepancy between the pre-training and fine-tuning stages. Moreover, to make the stage transition smoother, we train the NCT model using a gradual transition strategy, \emph{i.e.}, gradually transiting from using monolingual to bilingual dialogues. Extensive experiments on two language pairs demonstrate the effectiveness and superiority of our proposed training framework.
\end{abstract}

}

\maketitle

\IEEEdisplaynontitleabstractindextext

%
\IEEEpeerreviewmaketitle

\IEEEraisesectionheading{\section{Introduction}\label{sec:introduction}}

\begin{figure}[t]
\centering
\includegraphics[width=0.50\textwidth]{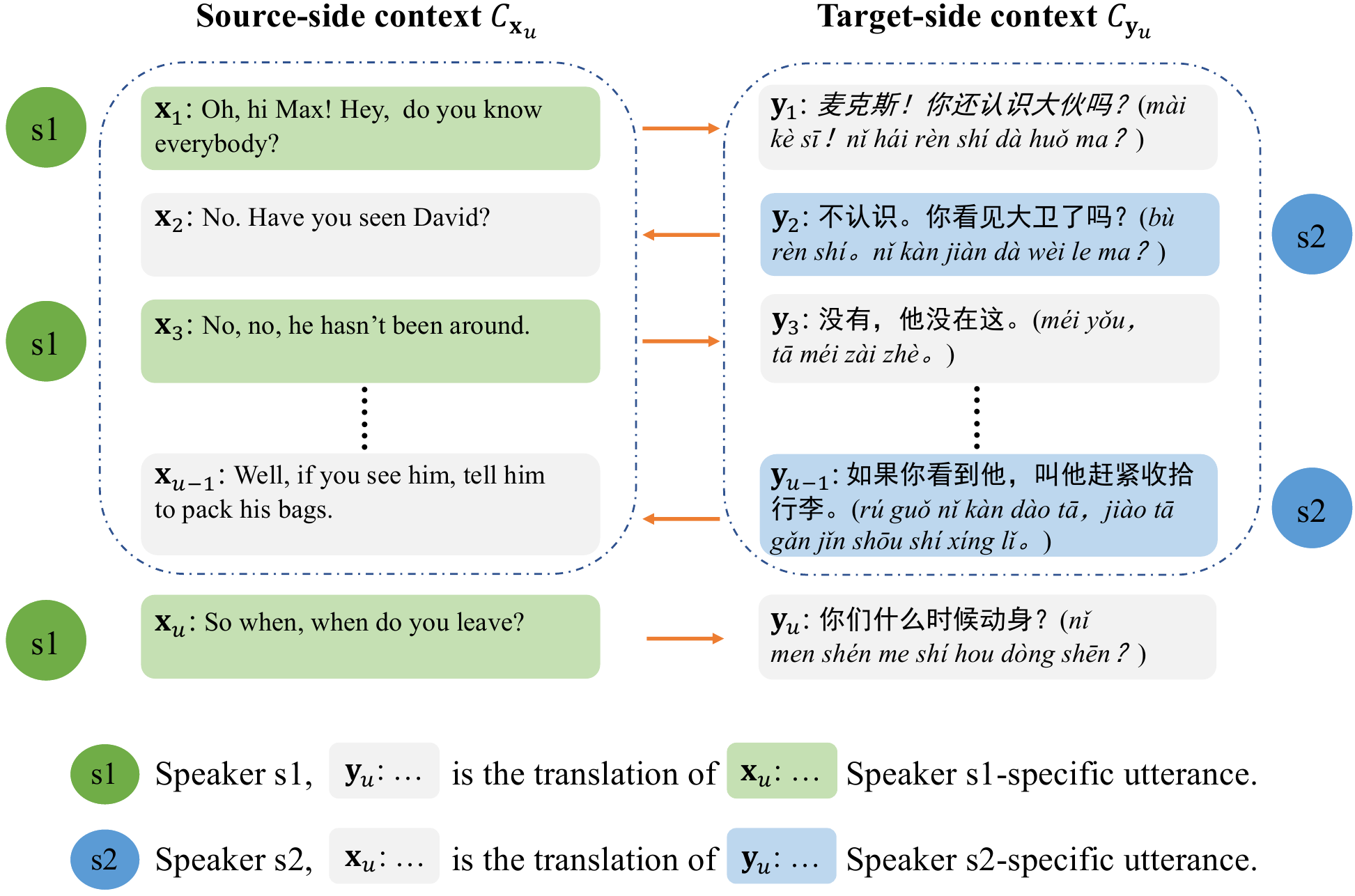}
\vspace{0pt}
\caption{An example of cross-lingual chat (En$\Leftrightarrow$Zh). The speaker s1-specific utterance $\mathbf{x}_{u}$ is being translated from English to Chinese with corresponding dialogue history context.}
\label{fig:ctx_case}
\vspace{-5pt}
\end{figure}

%
%
%
%
\vspace{10pt}
\IEEEPARstart{N}{eural} Chat Translation (NCT) is to translate a cross-lingual chat between speakers of different languages into utterances of their individual mother tongue. Fig.~\ref{fig:ctx_case} depicts an example of cross-lingual chat where one speaks in English and another in Chinese with their corresponding translations. With more international communication and cooperation all around the world, the chat translation task becomes more important and has broader applications in daily life. 

In this task, sentence-level Neural Machine Translation (NMT) models \cite{DBLP:conf/nips/SutskeverVL14,DBLP:journals/corr/BahdanauCB14,Vaswani:nips17} can be directly used to translate dialogue utterances sentence by sentence. In spite of its practicability, sentence-level NMT models often generate unsatisfactory translations due to ignoring the contextual information in dialogue history. To address this problem, many researches \cite{DBLP:conf/discomt/TiedemannS17,DBLP:conf/acl/HaffariM18,DBLP:conf/naacl/BawdenSBH18,DBLP:conf/emnlp/WerlenRPH18,DBLP:journals/tacl/TuLSZ18,DBLP:conf/acl/VoitaSST18,DBLP:conf/acl/VoitaST19a,DBLP:conf/emnlp/VoitaST19b,DBLP:conf/emnlp/WangTWS19,DBLP:conf/naacl/MarufMH19,DBLP:conf/acl/MaZZ20} adapt context-aware NMT models to make chat translation through their capability of incorporating dialogue history context. Generally, these methods adopt a pretrain-finetune paradigm, which first pre-train a sentence-level NMT model on a large-scale parallel corpus and then fine-tune it on the chat translation dataset in a context-aware way. However, they still can not obtain satisfactory results in the scenario of chat translation, mainly due to the following aspects of limitations: 1) 
The resource of bilingual chat translation corpus is usually limited, thus making an NCT model insufficiently trained to fully exploit dialogue context. 2) 
Conventional ways of incorporating dialogue context neglect to explicitly model its conversational properties such as dialogue coherence and speaker characteristic, resulting in incoherent and speaker-inconsistent translations. 3) 
The abrupt transition from sentence-level pre-training to context-aware fine-tuning breaks the consistency of model training, which hurts the potential performance of the final NCT model. Therefore, it is of great significance to train a better NCT models by resolving the above three aspects of limitations.

In this paper, we propose a \textbf{m}ulti-task \textbf{m}ulti-stage \textbf{t}ransitional (MMT) training framework where an NCT model is trained using the bilingual chat translation dataset and additional monolingual dialogues. Specifically, our proposed framework consists of three training stages, also following the pretrain-finetune paradigm. The first stage is still to pre-train the NCT model through sentence-level translation on the large-scale parallel corpus, resulting in the model $M_1$. At the second stage, using $M_1$ for model initialization, we continue to train the model through the previous sentence-level translation task along with two auxiliary dialogue-related tasks using additional monolingual dialogues, obtaining the model $M_2$. The auxiliary tasks are related to dialogue coherence and speaker characteristic, which are two important conversational properties of dialogue context. For the dialogue coherence, we design the task of \textit{\textbf{U}tterance} \textit{\textbf{D}iscrimination} (UD). The UD task is to judge whether an utterance and a given section of contextual utterances are within the same dialogue. For the speaker characteristic, we design the \textit{\textbf{S}peaker} \textit{\textbf{D}iscrimination} (SD) task. The SD task is to discriminate whether a given utterance and a piece of speaker-specific dialogue history contexts are spoken by the same speaker. Finally, at the last stage, initialized by $M_2$, the model is fine-tuned using a gradual transition strategy and eventually becomes a context-aware NCT model $M_3$. Concretely, the NCT model is trained through the objective comprised of chat translation, UD and SD tasks. During this process, we initially construct training samples for the two auxiliary tasks from additional monolingual dialogues and gradually transit to using bilingual dialogues.

The MMT training framework enhances the NCT model from the following aspects. Firstly, the relatively abundant monolingual dialogues function as a supplement to the scarce annotated bilingual dialogues, making the model more sufficiently trained to exploit dialogue context. Secondly, the UD and SD tasks are directly related to dialogue coherence and speaker characteristic, thus introducing the modelling of these two conversational properties into the NCT model. Thirdly, the second training stage serves as an intermediate phase that alleviates the discrepancy between sentence-level pre-training and context-aware fine-tuning. Particularly, it endows the model with the preliminary capability to capture dialogue context for the subsequent NCT training. It is notable that the two dialogue-related auxiliary tasks exist at both the second and third stages with different training data, which maintains the training consistency to some extent. Therefore, at the third stage, the NCT model can be more effectively fine-tuned to leverage dialogue context using the chat translation dataset with only a small number of annotated bilingual dialogues.

In essence, the major contributions of our paper are as follows:
\begin{itemize}
\setlength{\itemsep}{9pt}
\setlength{\parsep}{0pt}
\setlength{\parskip}{0pt}
\item In NCT, our work is the first attempt to use additional relatively abundant monolingual dialogues for training, which helps the model more sufficiently trained to capture dialogue context for chat translation.
\item We elaborately design two dialogue-related auxiliary tasks, namely utterance discrimination and speaker discrimination. This makes the model more capable of modelling dialogue coherence and speaker characteristic, which are two important conversational properties of dialogue context. 
\item We propose to alleviate the training discrepancy between pre-training and fine-tuning by introducing an intermediate stage (Stage 2) and adopting a gradual transition strategy for the context-aware fine-tuning (Stage 3). At the second stage, the model is simultaneously optimized with the two auxiliary tasks on the additional monolingual dialogues. Moreover, at the third stage, we train the NCT model by gradually transiting from using monolingual to bilingual dialogues, making the stage transition smoother. Thus, the NCT model can be more effectively fine-tuned on the small-scale bilingual chat translation dataset.
\item We will release the code of this work on Github https://
github.com/DeepLearnXMU.
\end{itemize}
\vspace{5pt}
The remainder of this paper is organized as follows. Section~\ref{sec:background} gives the NCT problem formalization, introduces the basic architecture of our NCT model and describes the conventional two-stage training including sentence-level pre-training and context-aware fine-tuning. Section~\ref{sec:mmt} elaborates our proposed MMT training framework. In Section~\ref{sec:experiment}, we report the experimental results and make in-depth analysis. Section~\ref{sec:related_work} summarizes the related work, mainly involving several existing studies on NCT and context-aware NMT models. Finally, in Section~\ref{sec:conclusion}, we draw the conclusions of this paper.


\begin{figure*}[ht]
\centering
\includegraphics[width=0.90\textwidth]{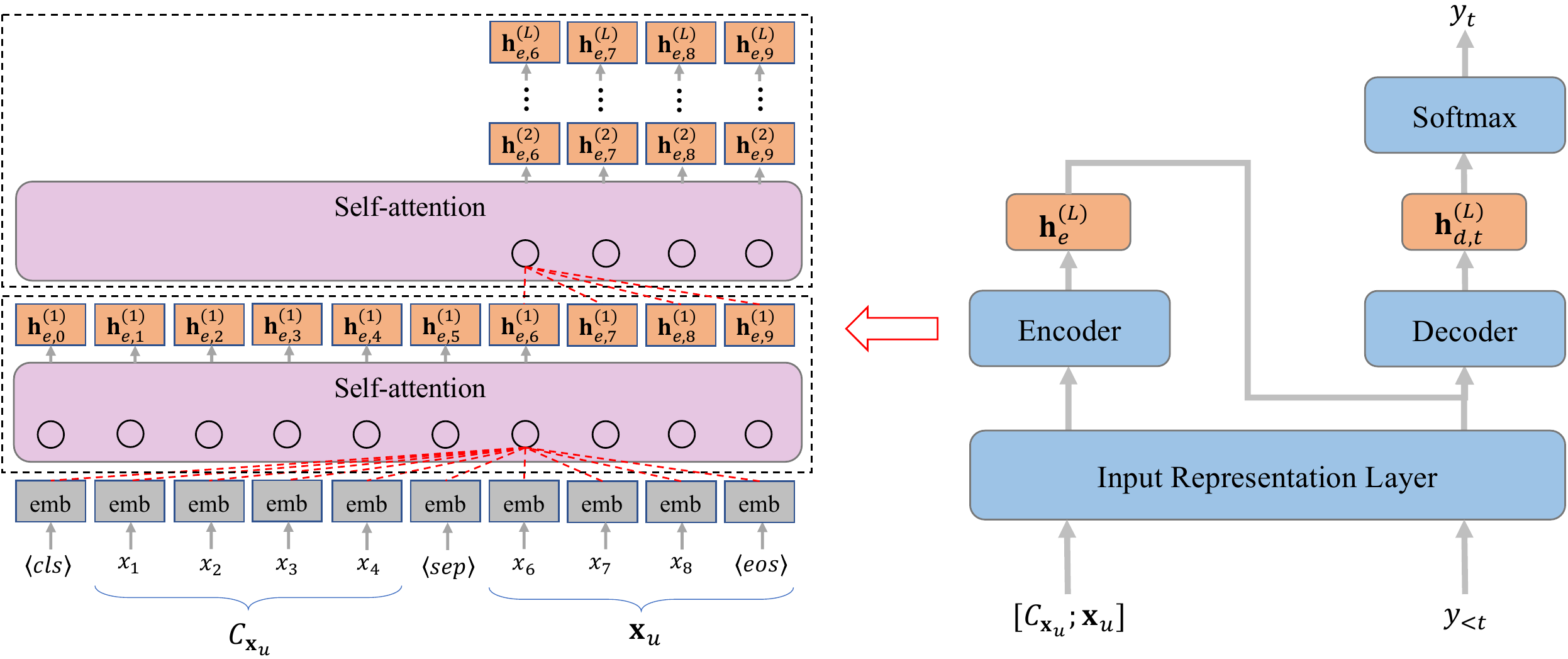}
\caption{The architecture of the Flat-NCT model used in this work. The left part depicts the attention mechanism inside Flat-NCT encoder. For illustration, we assume the input sequence $\mathcal{C}_{\mathbf{x}_u};\mathbf{x}_u$ is the concatenation of $\mathcal{C}_{\mathbf{x}_u}$$=$$x_1$,$x_2$,$x_3$,$x_4$ and $\mathbf{x}_u$$=$$x_6$,$x_7$,$x_8$,$\langle{eos}\rangle$ separated by a special token ``$\langle{sep}\rangle$''. Notably, words in $\mathcal{C}_{\mathbf{x}_u}$ can only be attended to by those in $\mathbf{x}_u$ at the first encoder layer. At the other encoder layers, $\mathcal{C}_{\mathbf{x}_u}$ is masked and the self-attention is only conducted within words of $\mathbf{x}_u$.}
\label{fig:nct_model}
\end{figure*}

\vspace{5pt}
\section{Background}
\label{sec:background}
In this section, we first give the NCT problem formalization (Section~\ref{subsec:prob_form}). Then, we describe the Flat-NCT model, which is the model architecture used in this work (Section~\ref{subsec:nct_model}). Finally, we introduce the dominant approach of training an NCT model, which consists of sentence-level pre-training (Section~\ref{subsec:sent_pretrain}) and context-aware fine-tuning (Section~\ref{subsec:context_ft}).
\begin{table}[]
	\centering
	\caption{Definitions of Different Dialogue History Contexts}
	\begin{tabular}{|c|c|c|}  
		\hline  
		& & \\[-6pt]
		Symbol & Definition & Meaning\\
		\hline
		& &  \\[-6pt]
		$\mathcal{C}_{\mathbf{x}_u}$ & $\mathbf{x}_1, \mathbf{x}_2, \mathbf{x}_3,..., \mathbf{x}_{u-1}$ & Source-side context of $\mathbf{x}_u$ \\
		\hline
		& &  \\[-6pt]
		$\mathcal{C}_{\mathbf{y}_u}$ & $\mathbf{y}_1, \mathbf{y}_2, \mathbf{y}_3,..., \mathbf{y}_{u-1}$ & Target-side context of $\mathbf{y}_u$ \\
		\hline
		& &  \\[-6pt]
		$\mathcal{C}_{\mathbf{x}_u}^{s1}$ & $\mathbf{x}_1, \mathbf{x}_3,..., \mathbf{x}_{u-2}$ & $s1$-specific context of $\mathbf{x}_u$ \\
		\hline
		& &  \\[-6pt]
		$\mathcal{C}_{\mathbf{x}_u}^{s2}$ & $\mathbf{x}_2, \mathbf{x}_4,..., \mathbf{x}_{u-1}$ & $s2$-specific context of $\mathbf{x}_u$ \\
		\hline
		& &  \\[-6pt]
		$\mathcal{C}_{\mathbf{y}_u}^{s1}$ & $\mathbf{y}_1, \mathbf{y}_3,..., \mathbf{y}_{u-2}$ & $s1$-specific context of $\mathbf{y}_u$ \\
		\hline
		& &  \\[-6pt]
		$\mathcal{C}_{\mathbf{y}_u}^{s2}$ & $\mathbf{y}_2, \mathbf{y}_4,..., \mathbf{y}_{u-1}$ & $s2$-specific context of $\mathbf{y}_u$ \\
		\hline
		& &  \\[-6pt]
		$\mathcal{C}_{\overline{\mathbf{x}}_u}$ & $\overline{\mathbf{x}}_1, \overline{\mathbf{x}}_2, \overline{\mathbf{x}}_3,..., \overline{\mathbf{x}}_{u-1}$ & Context of $\overline{\mathbf{x}}_u$ \\
		\hline
		& &  \\[-6pt]
		$\mathcal{C}_{\overline{\mathbf{y}}_u}$ & $\overline{\mathbf{y}}_1, \overline{\mathbf{y}}_2, \overline{\mathbf{y}}_3,..., \overline{\mathbf{y}}_{u-1}$ & Context of $\overline{\mathbf{y}}_u$ \\
		\hline
		& &  \\[-6pt]
		$\mathcal{C}_{\overline{\mathbf{x}}_u}^{s1}$ & $\overline{\mathbf{x}}_1, \overline{\mathbf{x}}_3,..., \overline{\mathbf{x}}_{u-2}$ & $s1$-specific context of $\overline{\mathbf{x}}_u$ \\
		\hline
		& & \\[-6pt]
		$\mathcal{C}_{\overline{\mathbf{x}}_u}^{s2}$ & $\overline{\mathbf{x}}_2, \overline{\mathbf{x}}_4,..., \overline{\mathbf{x}}_{u-1}$ & $s2$-specific context of $\overline{\mathbf{x}}_u$ \\
		\hline
		& & \\[-6pt]
		$\mathcal{C}_{\overline{\mathbf{y}}_u}^{s1}$ & $\overline{\mathbf{y}}_1, \overline{\mathbf{y}}_3,..., \overline{\mathbf{y}}_{u-2}$ & $s1$-specific context of $\overline{\mathbf{y}}_u$ \\
		\hline
		& & \\[-6pt]
		$\mathcal{C}_{\overline{\mathbf{y}}_u}^{s2}$ & $\overline{\mathbf{y}}_2, \overline{\mathbf{y}}_4,..., \overline{\mathbf{y}}_{u-1}$ & $s2$-specific context of $\overline{\mathbf{y}}_u$ \\
		\hline
	\end{tabular}
	\vspace{10pt}
	\begin{tablenotes}
	\item $\overline{\mathbf{x}}_u$ represents an utterance from the source-language monolingual dialogue $\overline{X}$ and $\overline{\mathbf{y}}_u$ is from the target-language monolingual dialogue $\overline{Y}$.
	\end{tablenotes}
	\label{tab:definition}
\end{table}

\subsection{Problem Formalization}
\label{subsec:prob_form}
In the scenario of this work, we denote the two speakers involved in a dialogue as $s1$ and $s2$. For a cross-lingual chat, as shown in the example in Fig.~\ref{fig:ctx_case}, the two speakers speak in the source and target language, respectively. We assume they have alternately given utterances in their own languages for $u$ turns, resulting in the source-language utterance sequence $X$$=$$\mathbf{x}_1, \mathbf{x}_2, \mathbf{x}_3, \mathbf{x}_4,..., \mathbf{x}_{u-1}, \mathbf{x}_u$ and the target-language utterance sequence $Y$$=$$\mathbf{y}_1, \mathbf{y}_2, \mathbf{y}_3, \mathbf{y}_4,...,\mathbf{y}_{u-1}, \mathbf{y}_u$. Notably, $X$ and $Y$ contain both the utterances originally spoken by one speaker and the translated utterances from the other speaker. Specifically, among these utterances, $\mathbf{x}_1, \mathbf{x}_3,..., \mathbf{x}_u$ are originally spoken by the source-language speaker $s1$ and $\mathbf{y}_1, \mathbf{y}_3,..., \mathbf{y}_u$ are the corresponding translations in the target language. Analogously, $\mathbf{y}_2, \mathbf{y}_4,..., \mathbf{y}_{u-1}$ are originally spoken by the target-language speaker $s2$ and $\mathbf{x}_2, \mathbf{x}_4,..., \mathbf{x}_{u-1}$ are the translated utterances in the source language. 

Besides the bilingual dialogues, our proposed training framework uses additional monolingual dialogues $D_{\overline{X}}$ of the source language and $D_{\overline{Y}}$ of the target language. Slightly different from the bilingual dialogue, the two speakers ($s1$ and $s2$) in a monolingual dialogue speak in the same language. We also assume a source-language monolingual dialogue $\overline{X}$$\in$$D_{\overline{X}}$ and a target-language monolingual $\overline{Y}$$\in$$D_{\overline{Y}}$ proceed to the $u$-th turn, resulting in $\overline{\mathbf{x}}_1, \overline{\mathbf{x}}_2, \overline{\mathbf{x}}_3, \overline{\mathbf{x}}_4,..., \overline{\mathbf{x}}_{u-1}, \overline{\mathbf{x}}_u$ and $\overline{\mathbf{y}}_1, \overline{\mathbf{y}}_2, \overline{\mathbf{y}}_3, \overline{\mathbf{y}}_4,..., \overline{\mathbf{y}}_{u-1}, \overline{\mathbf{y}}_u$, respectively.

Then, we give the necessary definitions in the remainder of this paper. For clarity, we list all definitions\footnote{For each item of \{$C_{\mathbf{x}_u}$, $C_{\mathbf{y}_u}$, $C_{\mathbf{x}_u}^{s1}$, $C_{\mathbf{x}_u}^{s2}$, $C_{\mathbf{y}_u}^{s1}$, $C_{\mathbf{y}_u}^{s2}$, $C_{\overline{\mathbf{x}}_u}$, $C_{\overline{\mathbf{y}}_u}$, $C_{\overline{\mathbf{x}}_u}^{s1}$, $C_{\overline{\mathbf{x}}_u}^{s2}$, $C_{\overline{\mathbf{y}}_u}^{s1}$, $C_{\overline{\mathbf{y}}_u}^{s2}$\}, taking $C_{\mathbf{x}_u}$ for instance, we prepend a special token `[cls]' to it and use another special token `[sep]' to delimit its included utterances, as implemented in~\cite{DBLP:conf/naacl/DevlinCLT19}.} in Table~\ref{tab:definition}. For a bilingual dialogue, we define the dialogue history context of $\mathbf{x}_u$ on the source side as $\mathcal{C}_{\mathbf{x}_u}$=$\mathbf{x}_1, \mathbf{x}_2, \mathbf{x}_3,..., \mathbf{x}_{u-1}$ and that of $\mathbf{y}_u$ on the target side as $\mathcal{C}_{\mathbf{y}_u}$=$\mathbf{y}_1, \mathbf{y}_2, \mathbf{y}_3,..., \mathbf{y}_{u-1}$. According to original speakers, on the source side, we define the speaker $s1$-specific dialogue history context of $\mathbf{x}_u$ as the partial sequence of its preceding utterances $\mathcal{C}_{\mathbf{x}_u}^{s1}$=$\mathbf{x}_1, \mathbf{x}_3,..., \mathbf{x}_{u-2}$ and the speaker $s2$-specific dialogue history context of $\mathbf{x}_u$ as $\mathcal{C}_{\mathbf{x}_u}^{s2}$=$\mathbf{x}_2, \mathbf{x}_4,..., \mathbf{x}_{u-1}$. On the target side, $\mathcal{C}_{\mathbf{y}_u}^{s1}$=$\mathbf{y}_1, \mathbf{y}_3,..., \mathbf{y}_{u-2}$ and $\mathcal{C}_{\mathbf{y}_u}^{s2}$=$\mathbf{y}_2, \mathbf{y}_4,..., \mathbf{y}_{u-1}$ denote the speaker $s1$-specific and $s2$-specific dialogue history contexts of $\mathbf{y}_u$, respectively. When it comes to a monolingual dialogue, we also formalize different types of dialogue history contexts \{$C_{\overline{\mathbf{x}}_u}$, $C_{\overline{\mathbf{y}}_u}$, $C_{\overline{\mathbf{x}}_u}^{s1}$, $C_{\overline{\mathbf{x}}_u}^{s2}$, $C_{\overline{\mathbf{y}}_u}^{s1}$, $C_{\overline{\mathbf{y}}_u}^{s2}$\} in a similar way.


\subsection{The NCT model}
\label{subsec:nct_model}
We use the Flat-Transformer introduced in~\cite{DBLP:conf/acl/MaZZ20} as our basic NCT model, which we denote as Flat-NCT. Figure~\ref{fig:nct_model} shows the architecture of the Flat-NCT, mainly including \emph{input representation layer}, \emph{encoder} and \emph{decoder}.

\subsubsection{Input Representation Layer}
For each utterance $\mathbf{x}_u$$=$$x_{1}, x_{2},\!\cdots\!,x_{|\mathbf{x}_u|}$ to be translated, $[\mathcal{C}_{\mathbf{x}_u};\mathbf{x}_u]$ is fed into the NCT model as input, where $[;]$ denotes the concatenation. Different from the conventional embedding layer that only includes word embedding $\mathbf{WE}$ and position embedding $\mathbf{PE}$, we additionally add a speaker embedding $\mathbf{SE}$ and a turn embedding $\mathbf{TE}$. The final embedding $\mathbf{B}(x_{i})$ of each input word $x_{i}$ can be written as
\begin{equation}\label{input_embed}
\setlength{\abovedisplayskip}{10pt}
\setlength{\belowdisplayskip}{10pt}
\mathbf{B}(x_{i}) = \mathbf{WE}({x_{i}}) + \mathbf{PE}({x_{i}}) + \mathbf{SE}({x_{i}}) + \mathbf{TE}({x_{i}}),
\end{equation}
where $\mathbf{WE}\in{\mathbb{R}^{|V|\times{d}}}$, $\mathbf{SE}\in{\mathbb{R}^{2\times{d}}}$ and $\mathbf{TE}\in{\mathbb{R}^{|U|\times{d}}}$. Here, $|V|$, $|U|$ and $d$ denote the size of shared vocabulary, maximum dialogue turns, and the hidden size, respectively.

\subsubsection{Encoder}
The encoder of our NCT model has $L$ identical layers, each of which is composed of a self-attention ($\mathrm{SelfAtt}$) sub-layer and a feed-forward network ($\mathrm{FFN}$) sub-layer.\footnote{The layer normalization is omitted for simplicity.} Let $\mathbf{h}^{(l)}_e$ denote the hidden states of the $l$-th encoder layer, it is calculated using the following equations:
\begin{equation}
\setlength{\abovedisplayskip}{10pt}
\setlength{\belowdisplayskip}{10pt}
\begin{split}
    \mathbf{z}^{(l)}_e &= \mathrm{SelfAtt}(\mathbf{h}^{(l-1)}_e) + \mathbf{h}^{(l-1)}_e,\\
    \mathbf{h}^{(l)}_e &= \mathrm{FFN}(\mathbf{z}^{(l)}_e) + \mathbf{z}^{(l)}_e,
\end{split}
\end{equation}
where $\mathbf{h}^{(0)}_e$ is initialized as the embedding of input words.
Particularly, words in $\mathcal{C}_{\mathbf{x}_u}$ can only be attended to by those in $\mathbf{x}_u$ at the first encoder layer while $\mathcal{C}_{\mathbf{x}_u}$ is masked at the other layers, as implemented in~\cite{DBLP:conf/acl/MaZZ20}.

\subsubsection{Decoder}
The decoder also consists of $L$ identical layers, each of which additionally has a cross-attention ($\mathrm{CrossAtt}$) sub-layer compared to the encoder. Let $\mathbf{h}^{(l)}_d$ denote the hidden states of the $l$-th decoder layer, it is computed as
\begin{equation}
\setlength{\abovedisplayskip}{10pt}
\setlength{\belowdisplayskip}{10pt}
\begin{split}
    \mathbf{z}^{(l)}_d &= \mathrm{SelfAtt}(\mathbf{h}^{(l-1)}_d) + \mathbf{h}^{(l-1)}_d,\\
    \mathbf{c}^{(l)}_d &= \mathrm{CrossAtt}(\mathbf{z}^{(l)}_d, \mathbf{h}_e^{(L)}) + \mathbf{z}^{(l)}_d,\\
    \mathbf{h}^{(l)}_d &= \mathrm{FFN}(\mathbf{c}^{(l)}_d) + \mathbf{c}^{(l)}_d,
\end{split}
\end{equation}
where $\mathbf{h}^{(L)}_e$ corresponds to the top-layer encoder hidden states.

At each decoding time step $t$, the $t$-th decoder hidden state $\mathbf{h}^{(L)}_{d,t}$ is fed into a linear transformation layer and a softmax layer to predict the probability distribution of the next target token:
\begin{equation}
\setlength{\abovedisplayskip}{10pt}
\setlength{\belowdisplayskip}{10pt}
    p(y_{t}|y_{<t}, \mathbf{x}_u, \mathcal{C}_{\mathbf{x}_u}) = \mathrm{Softmax}(\mathbf{W}_o\mathbf{h}^{(L)}_{d,t}+\mathbf{b}_o),
\end{equation}
where $\mathbf{W}_o \in \mathbb{R}^{|V|\times d}$ and $\mathbf{b}_o \in \mathbb{R}^{|V|}$ are trainable parameters.

\subsection{Two-stage Training}
\subsubsection{Sentence-level Pre-training}
\label{subsec:sent_pretrain}
At this stage, the NCT model is pre-trained on a large-scale parallel corpus $D_{sent}$ in the way of a vanilla sentence-level translation. For each parallel sentence pair $(\mathbf{x}, \mathbf{y})$ $\in$ $D_{sent}$, taking $\mathbf{x}$ as input, the model is optimized through the following objective:
\begin{equation}
\setlength{\abovedisplayskip}{10pt}
\setlength{\belowdisplayskip}{10pt}
\label{eq:nmt}
    \mathcal{L}_{sent}(\theta_{nct}) = \sum_{t=1}^{|\mathbf{y}|}\mathrm{log}(p(y_{t}|\mathbf{x}, y_{<t})),
\end{equation}
where $\theta_{nct}$ is the parameters of the NCT model, $\mathbf{y}$$=$$y_1,y_2,\cdots,y_{|\mathbf{y}|}$ is the target translation, $y_{t}$ is the $t$-th word of $\mathbf{y}$ and $y_{<t}$ denotes the partial sequence $y_{1},\cdots,y_{t-1}$ of target words preceding $y_{t}$. 

\subsubsection{Context-aware Fine-tuning}
\label{subsec:context_ft}
After the sentence-level pre-training, the model is then fine-tuned using the bilingual chat translation dataset $D_{bct}$ in a context-aware way. Concretely, given a piece of $U$-turn parallel bilingual dialogue utterances $(X,Y)$ $\in$ ${D_{bct}}$, where $X$$=$$\mathbf{x}_1,\mathbf{x}_2,\cdots,\mathbf{x}_U$ is in the source language while $Y$$=$$\mathbf{y}_1,\mathbf{y}_2,\cdots,\mathbf{y}_U$ is in the target language,\footnote{Note that $X$ contains both the utterances originally spoken by the source-language speaker and the translations of those originally spoken by the other speaker of the target language, which is the same for $Y$.} the training objective at this stage can be formalized as
\begin{equation}
\label{eq:ca-finetune}
\setlength{\abovedisplayskip}{10pt}
\setlength{\belowdisplayskip}{10pt}
    \mathcal{L}_{nct}(\theta_{nct}) = -\sum_{u=1}^{U}\mathrm{log}(p(\mathbf{y}_u|\mathbf{x}_u, \mathbf{x}_{<u}, \mathbf{y}_{<u})),
\end{equation}
where $\mathbf{x}_{<u}$ and $\mathbf{y}_{<u}$ are the preceding utterance sequences of the $u$-th source-language utterance $\mathbf{x}_u$ and the $u$-th target-language utterance $\mathbf{y}_u$, respectively. More specifically, $p(\mathbf{y}_u|\mathbf{x}_u, \mathbf{x}_{<u}, \mathbf{y}_{<u})$ is calculated as
\begin{equation}
\label{eq:word-level_prob}
\setlength{\abovedisplayskip}{10pt}
\setlength{\belowdisplayskip}{10pt}
    p(\mathbf{y}_u|\mathbf{x}_u, \mathbf{x}_{<u}, \mathbf{y}_{<u}) = \prod_{t=1}^{|\mathbf{y}_u|} p(y_{t}|y_{<t}, \mathbf{x}_u, \mathbf{x}_{<u}, \mathbf{y}_{<u}),
\end{equation}
where $y_t$ is the $t$-th target word in $\mathbf{y}_u$ and $y_{<t}$ denotes the preceding tokens $y_{1},y_{2},\cdots,y_{t-1}$ before the $t$-th time step.

\begin{figure*}[ht]
\centering
  \includegraphics[width = 1.0\textwidth]{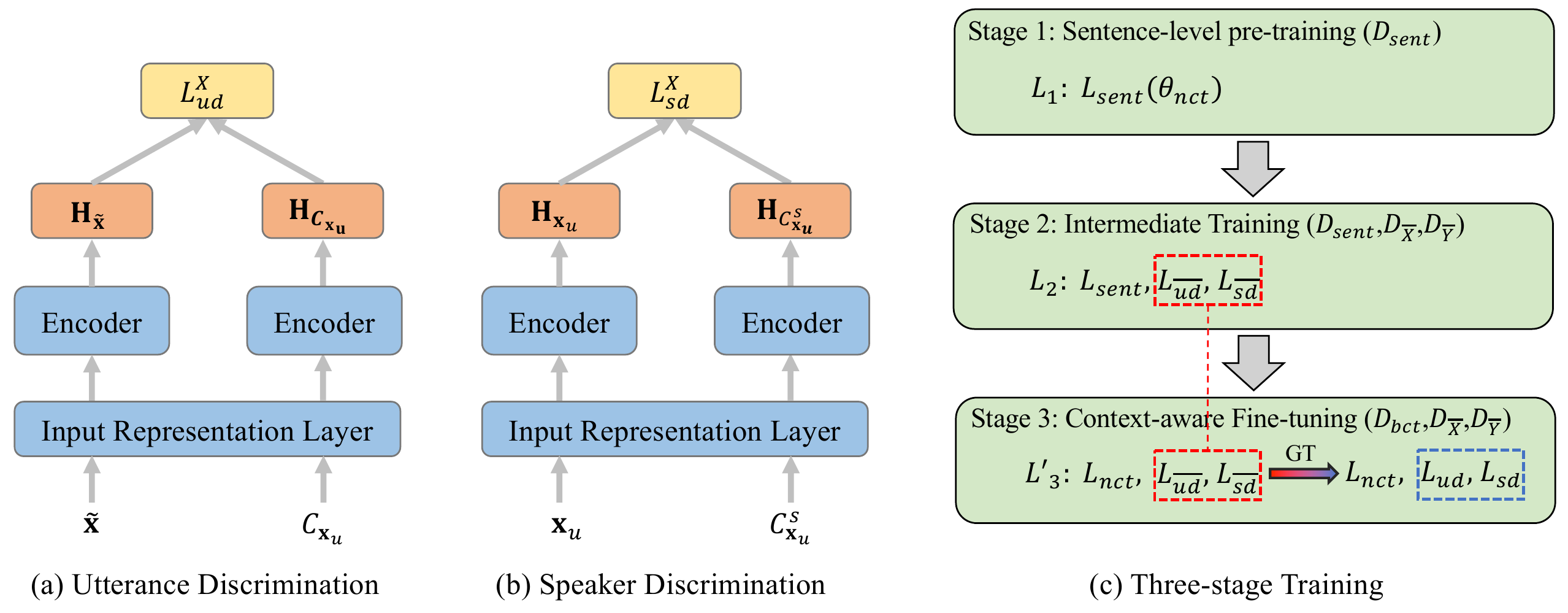}
\caption{Overview of the auxiliary tasks and the MMT training framework. To show the two auxiliary tasks, we just take the source-language dialogue $X$$=$$\mathbf{x}_1, \mathbf{x}_2, \mathbf{x}_3, \mathbf{x}_4,..., \mathbf{x}_{u-1}, \mathbf{x}_u$ for instance, which can be analogously generalized to other types of dialogues ($Y$, $\overline{X}$ and $\overline{Y}$). (a): The utterance discrimination (UD) task. (b): The speaker discrimination (SD) task. (c): The three training stages of our proposed framework. Note that the NCT encoder is shared across the chat translation and the two auxiliary tasks.}
\label{fig:ud_sd_train}
\vspace{0pt}
\end{figure*}

\section{Multi-task Multi-stage Transitional Training Framework}
\label{sec:mmt}
In this section, we give a detailed description of our proposed multi-task multi-stage transitional (MMT) training framework for NCT, which aims to improve the NCT model with dialogue-related auxiliary tasks using additional monolingual dialogues. In the following subsections, we first introduce the two proposed dialogue-related auxiliary tasks (Section~\ref{subsec:aux_task}) in detail. Then, we elaborate the procedures of our proposed training framework (Section~\ref{subsec:model_training}).

\subsection{Auxiliary Tasks}
\label{subsec:aux_task}
In our proposed training framework, we elaborately design two auxiliary tasks that are related to two important conversational properties of dialogue context, namely dialogue coherence and speaker characteristic. The first task for dialogue coherence is utterance discrimination (UD) and the second for speaker characteristic is speaker discrimination (SD). Together with the main chat translation task, the NCT model can be enhanced to generate more coherent and speaker-consistent translations through multi-task learning. In the following subsections, in order to clearly describe the two auxiliary tasks, we just take a source-language dialogue $X$$=$$\mathbf{x}_1, \mathbf{x}_2, \mathbf{x}_3, \mathbf{x}_4,..., \mathbf{x}_{u-1}, \mathbf{x}_u$ for instance, which can be generalized to other types of dialogues ($Y$, $\overline{X}$ and $\overline{Y}$).

\subsubsection{Utterance Discrimination (UD)}
\label{subsubsec:task_ud}
A series of previous studies~\cite{DBLP:conf/coling/KuangXLZ18,DBLP:conf/emnlp/WangFWZ19,DBLP:conf/aaai/XiongH0W19,DBLP:conf/ijcai/Wang019b,DBLP:conf/emnlp/HuangYQLL20} have indicated that the modelling of global contextual coherence can lead to more coherent generated text. From this perspective, we design the task of UD to introduce the modelling of dialogue coherence into the NCT model. 

As shown in Fig.~\ref{fig:ud_sd_train}(a), our UD task aims to distinguish whether an utterance and a given section of contextual utterances are within the same dialogue. To this end, we construct positive and negative training samples from the monolingual and bilingual dialogues, where a training sample ($\mathcal{C}_{\mathbf{x}_{u}}$, $\widetilde{\mathbf{x}}$) contains a section of dialogue history context $\mathcal{C}_{\mathbf{x}_{u}}$ and a selected utterance $\widetilde{\mathbf{x}}$ with the label $\ell^{X}_{ud}$. For a positive sample with label $\ell^{X}_{ud}=1$, $\widetilde{\mathbf{x}}$ is exactly $\mathbf{x}_{u}$, while for a negative sample with label $\ell^{X}_{ud}=0$, $\widetilde{\mathbf{x}}$ is a randomly selected utterance from any other irrelevant dialogue. Formally, the training objective of UD is defined as follows:
\begin{equation}
\label{eq:task_ud}
\setlength{\abovedisplayskip}{10pt}
\setlength{\belowdisplayskip}{10pt}
    \mathcal{L}_{ud}^{X}(\theta_{nct},\theta_{ud})  =-\mathrm{log}(p(\hat{\ell}^{X}_{ud}=\ell^{X}_{ud}|\mathcal{C}_{\mathbf{x}_{u}}, \widetilde{\mathbf{x}})),
\end{equation}
where $\theta_{nct}$ and $\theta_{ud}$ are the trainable parameters of the NCT model and UD classifier, respectively.

To estimate the probability in Eq.~\ref{eq:task_ud}, we first obtain the representations $\mathbf{H}_{\widetilde{\mathbf{x}}}$ of the utterance $\widetilde{\mathbf{x}}$ and $\mathbf{H}_{\mathcal{C}_{\mathbf{x}_u}}$ of the dialogue history context $\mathcal{C}_{\mathbf{x}_u}$ using the NCT encoder. Specifically, $\mathbf{H}_{\widetilde{\mathbf{x}}}$ is calculated as $\frac{1}{|\widetilde{\mathbf{x}}|}\sum_{i=1}^{|\widetilde{\mathbf{x}}|}\mathbf{h}^{(L)}_{e,i}$ while $\mathbf{H}_{\mathcal{C}_{\mathbf{x}_u}}$ is defined as the encoder hidden state $\mathbf{h}^{(L)}_{e,0}$ of the prepended special token `[cls]' in $\mathcal{C}_{\mathbf{x}_u}$. Then, the concatenation of $\mathbf{H}_{\mathbf{x}}$ and $\mathbf{H}_{\mathcal{C}_{\mathbf{x}_u}}$ is fed into a binary UD classifier, which is an extra fully-connected layer on top of the NCT encoder:
\begin{equation}
\setlength{\abovedisplayskip}{10pt}
\setlength{\belowdisplayskip}{10pt}
\begin{split}
    p(\hat{\ell}^{X}_{ud}&=1|\mathcal{C}_{\mathbf{x}_u}, \widetilde{\mathbf{x}})=\mathrm{sigmoid}(\mathbf{W}_{ud}[\mathbf{H}_{\widetilde{\mathbf{x}}}; \mathbf{H}_{\mathcal{C}_{\mathbf{x}_u}}]),\\
    p(\hat{\ell}^{X}_{ud}&=0|\mathcal{C}_{\mathbf{x}_u}, \widetilde{\mathbf{x}})=1-p(\hat{\ell}^{X}_{ud}=1|\mathcal{C}_{\mathbf{x}_u}, \widetilde{\mathbf{x}}),
\end{split}
\end{equation}
where $\mathbf{W}_{ud}$ is the trainable parameter matrix of the UD classifier and the bias term is omitted for simplicity.

\subsubsection{Speaker Discrimination (SD)}
\label{subsubsec:task_sd}
Generally, a dialogue may involve speakers with different characteristics, which is a salient conversational property. Therefore, we design the SD task to incorporate the modelling of speaking style into the NCT model, making the translated utterance more speaker-consistent.

As shown in Fig.~\ref{fig:ud_sd_train}(b), the SD task is to discriminate whether a given utterance and a piece of speaker-specific dialogue history contexts are spoken by the same speaker. Similarly, we construct positive and negative training samples from the monolingual and bilingual dialogues. Specifically, an SD training sample ($\mathcal{C}_{\mathbf{x}_u}^{s}$, $\mathbf{x}_u$) is comprised of the speaker $s$-specific dialogue history context ($s$ $\in$ $\{s1,s2\}$) and the utterance $\mathbf{x}_u$ with the corresponding label $\ell^{X}_{sd}$. For a positive sample with label $\ell^{X}_{sd}=1$, the dialogue history context is specific to the speaker $s1$ ($\mathbf{x}_u$ is spoken by $s1$), while for a negative sample with label $\ell^{X}_{sd}=0$, it is specific to the other speaker $s2$. Formally, the training objective of SD is defined as follows:
\begin{equation}
\label{eq:task_sd}
\setlength{\abovedisplayskip}{10pt}
\setlength{\belowdisplayskip}{10pt}
    \mathcal{L}_{sd}^{X}(\theta_{nct},\theta_{sd})  =-\mathrm{log}(p(\hat{\ell}^{X}_{sd}=\ell^{X}_{sd}|\mathcal{C}_{\mathbf{x}_{u}}^{s}, \mathbf{x}_u)),
\end{equation}
where $\theta_{nct}$ and $\theta_{sd}$ are the trainable parameters of the NCT model and SD classifier, respectively.

Analogously, we use the NCT encoder to obtain the representations $\mathbf{H}_{\mathbf{x}_u}$ of $\mathbf{x}_u$ and $\mathbf{H}_{\mathcal{C}_{\mathbf{x}_u}^{s}}$ of $\mathcal{C}_{\mathbf{x}_u}^{s}$, where $\mathbf{H}_{\mathbf{x}_u}$$=$$\frac{1}{|\mathbf{x}_u|}\sum_{i=1}^{|\mathbf{x}_u|}\mathbf{h}^{(L)}_{e,i}$ and the $\mathbf{h}^{(L)}_{e,0}$ of $\mathcal{C}_{\mathbf{x}_u}^{s}$ is used as $\mathbf{H}_{\mathcal{C}_{\mathbf{x}_u}^{s}}$. Then, to estimate the probability in Eq.~\ref{eq:task_sd}, the concatenation of $\mathbf{H}_{\mathbf{x}_u}$ and $\mathbf{H}_{\mathcal{C}_{\mathbf{x}_u}^{s}}$ is fed into a binary SD classifier, which is another fully-connected layer on top of the NCT encoder:
\begin{equation}
\setlength{\abovedisplayskip}{10pt}
\setlength{\belowdisplayskip}{10pt}
\begin{split}
    p(\hat{\ell}^{X}_{sd}=1|\mathcal{C}_{\mathbf{x}_u}^{s}, \mathbf{x}_u)&=\mathrm{sigmoid}(\mathbf{W}_{sd}[\mathbf{H}_{\mathbf{x}_u}; \mathbf{H}_{\mathcal{C}_{\mathbf{x}_u}^{s}}]),\\
    p(\hat{\ell}^{X}_{sd}=0|\mathcal{C}_{\mathbf{x}_u}^{s}, \mathbf{x}_u)&=1-p(\hat{\ell}^{X}_{sd}=1|\mathcal{C}_{\mathbf{x}_u}^{s}, \mathbf{x}_u),
\end{split}
\end{equation}
where $\mathbf{W}_{sd}$ is the trainable parameter matrix of the SD classifier and the bias term is omitted for simplicity.

\subsection{Three-stage Training}
\label{subsec:model_training}
Then, we elaborate the procedures of our proposed MMT training framework. The training totally consists of three stages: 1) sentence-level pre-training on large-scale parallel corpus; 2) intermediate training with auxiliary tasks using additional monolingual dialogues; 3) context-aware fine-tuning with gradual transition. During inference, the auxiliary tasks (UD and SD) are not involved and only the NCT model ($\theta_{nct}$) is used to conduct chat translation.
\subsubsection{Stage 1: Sentence-level Pre-training on Large-scale Parallel Corpus}
\label{subsec:stage1}
As described in Section~\ref{subsec:sent_pretrain}, the first stage is to grant the NCT model the basic capability of translating sentences.
Given the large-scale parallel corpus $D_{sent}$, we pre-train the model $M_1$ using the same training objective as Eq.~\ref{eq:nmt}, \emph{i.e.}, $\mathcal{L}_{1}$$=$$\mathcal{L}_{sent}(\theta_{nct})$.



\subsubsection{Stage 2: Intermediate Training with Auxiliary Tasks using Additional Monolingual Dialogues}
Under our proposed training framework, the second stage serves as an intermediate phase that involves additional monolingual dialogues, endowing the original context-agnostic model with the preliminary capability of capturing dialogue context. Using the pre-trained $M_1$ for model initialization, we continue to train the model through the previous sentence-level translation along with the two designed auxiliary tasks (UD and SD) using additional monolingual dialogues, obtaining the model $M_2$. 

Concretely, for UD and SD tasks, we construct training instances from $\overline{X}$ $\in$ $D_{\overline{X}}$ and $\overline{Y}$ $\in$ $D_{\overline{Y}}$ in the way described in Section~\ref{subsubsec:task_ud} and Section~\ref{subsubsec:task_sd}. Together with the sentence-level translation, the training objective at this stage can be written as
\begin{equation}
\label{eq:stage2}
\setlength{\abovedisplayskip}{10pt}
\setlength{\belowdisplayskip}{5pt}
     \mathcal{L}_{2}=\mathcal{L}_{sent}+\alpha_{1}\mathcal{L}_{\overline{ud}}+\beta_{1}\mathcal{L}_{\overline{sd}}, 
\end{equation}
\begin{equation}
\setlength{\abovedisplayskip}{5pt}
\setlength{\belowdisplayskip}{10pt}
\begin{split}
    \nonumber
    \text{where}\quad\, &\mathcal{L}_{\overline{ud}}=\mathcal{L}_{ud}^{\overline{X}}(\theta_{nct},\theta_{ud}) + \mathcal{L}_{ud}^{\overline{Y}}(\theta_{nct},\theta_{ud}), \\ 
    &\mathcal{L}_{\overline{sd}}=\mathcal{L}_{sd}^{\overline{X}}(\theta_{nct},\theta_{sd}) + \mathcal{L}_{sd}^{\overline{Y}}(\theta_{nct},\theta_{sd}),
\end{split}
\end{equation}
and $\alpha_{1}$ and $\beta_{1}$ are balancing hyper-parameters for the trade-off between $\mathcal{L}_{sent}$ and the other auxiliary objectives. Here, as similarly defined in Eq.~\ref{eq:task_ud} and Eq.~\ref{eq:task_sd}, $\mathcal{L}_{ud}^{\overline{X}}(\theta_{nct},\theta_{ud})$ and $\mathcal{L}_{ud}^{\overline{Y}}(\theta_{nct},\theta_{ud})$ represent the training objectives of the UD task on the source-language monolingual dialogue $\overline{X}$ and target-language monolingual dialogue $\overline{Y}$ respectively, which is analogous to $\mathcal{L}_{sd}^{\overline{X}}(\theta_{nct},\theta_{ud})$ and $\mathcal{L}_{sd}^{\overline{Y}}(\theta_{nct},\theta_{ud})$ of the SD task.

In this way, the tasks of UD and SD introduce the modelling of dialogue coherence and speaker characteristic into the sentence-level translation model. Meanwhile, we still use the objective $\mathcal{L}_{sent}$ so as to avoid undermining the pre-trained translation capability of the model, providing a better starting point for the subsequent NCT fine-tuning. 
\vspace{0pt}
\subsubsection{Stage 3: Context-aware Fine-tuning with Gradual Transition}
\label{subsubsec:stage3}
Using the bilingual chat translation dataset $D_{bct}$, the third stage is to obtain the final NCT model $M_{3}$ through context-aware fine-tuning, where the two auxiliary tasks (UD and SD) are still involved. Particularly, different from the second stage, we construct the training instances of UD and SD tasks from $X$ and $Y$. 

Given a bilingual dialogue pair $(X,Y)$ $\in$ $D_{bct}$, we optimize the model (initialized by $M_{2}$) through the following objective:
\begin{equation}
\label{eq:stage3}
\setlength{\abovedisplayskip}{10pt}
\setlength{\belowdisplayskip}{10pt}
    \mathcal{L}_{3}=\mathcal{L}_{nct}+\alpha_{2}\mathcal{L}_{ud}+\beta_{2}\mathcal{L}_{sd},
\end{equation}
\begin{equation}
\setlength{\abovedisplayskip}{10pt}
\setlength{\belowdisplayskip}{10pt}
\begin{split}
    \nonumber
    \text{where}\quad\, &\mathcal{L}_{ud}=\mathcal{L}_{ud}^{X}(\theta_{nct},\theta_{ud}) + \mathcal{L}_{ud}^{Y}(\theta_{nct},\theta_{ud}), \\ 
    &\mathcal{L}_{sd}=\mathcal{L}_{sd}^{X}(\theta_{nct},\theta_{sd}) + \mathcal{L}_{sd}^{Y}(\theta_{nct},\theta_{sd}),
\end{split}
\end{equation}
and $\alpha_{2}$ and $\beta_{2}$ are also the hyper-parameters controlling the balance between $\mathcal{L}_{nct}$ and the other auxiliary objectives analogously defined as in Eq.~\ref{eq:task_ud} or Eq.~\ref{eq:task_sd}. Notably, under our proposed training framework, UD and SD tasks exist both at the second and the third stages, which can benefit the NCT model in the following two aspects. On the one hand, the two auxiliary tasks maintain the training consistency, making the transition from sentence-level pre-training to context-aware fine-tuning smoother. On the other hand, because the model has acquired the preliminary capability of capturing dialogue context obtained at the second stage, it can be more effectively fine-tuned on $D_{bct}$ with only a small number of annotated bilingual dialogues. 

However, although the above strategy maintains the training consistency to some extent, the transition of training stage is still abrupt because the NCT model is trained with the two auxiliary tasks using totally different data at the second and third stages. To further alleviate the training discrepancy, we propose to train the NCT model by gradually transiting from using monolingual to bilingual dialogues. Specifically, we keep on using the additional monolingual dialogues ($\overline{X}$ and $\overline{Y}$) to accomplish a smoother transition of training stages. Therefore, the training objective of this stage can be formalized as
\begin{equation}
\label{eq:stage3_new}
\setlength{\abovedisplayskip}{10pt}
\setlength{\belowdisplayskip}{10pt}
\begin{split}
    \mathcal{L'}_{3}=\mathcal{L}_{nct}&+\lambda(\alpha_{2}\mathcal{L}_{ud} +\beta_{2}\mathcal{L}_{sd}) \\
    &+(1-\lambda)(\alpha_{1}\mathcal{L}_{\overline{ud}}+\beta_{1}\mathcal{L}_{\overline{sd}}),
\end{split}
\end{equation}
where $\lambda$$=$$n/N$ denotes the coefficient controlling the balance between monolingual and bilingual dialogues with $n$ being the current training step at the third stage and $N$ being the maximum steps of this stage. Note that $\alpha_{1}$ and $\beta_{1}$ are kept fixed as the values in Eq.~\ref{eq:stage2}. Considering that the additional monolingual dialogues are much more than the available annotated bilingual dialogues, they can function as a supplement to the scarce annotated bilingual dialogues, helping the model learn to better exploit dialogue context.

\section{Experiments}
\label{sec:experiment}
To investigate the effectiveness of our proposed training framework, we conducted experiments on {English}$\Leftrightarrow${German} ({En}$\Leftrightarrow${De}) and {English}$\Leftrightarrow${Chinese} ({En}$\Leftrightarrow${Zh}) chat translation datasets.

\begin{table}[t]
\renewcommand\arraystretch{1.5}
\caption{Dataset Statistics}
\centering
\setlength{\tabcolsep}{0.8mm}{
\begin{tabular}{c|c|c|c}
\toprule
\hline
\vspace{-2pt}
\textbf{Dataset}/\textbf{Split} \quad & Train & \;Valid \quad & \;Test \quad \\
\hline
WMT20 (En$\Leftrightarrow$De) & 45,541,367  & -  & - \\
\hline
WMT20 (En$\Leftrightarrow$Zh) & 22,244,006  & -  & - \\
\hline
Taskmaster-1 (En) & 153,774  & - & - \\
\hline
BConTrasT (En$\Rightarrow$De) &7,629 &1,040 &1,133 \\
\hline
BConTrasT (De$\Rightarrow$En)&6,216 &862 &967 \\
\hline
BMELD (En$\Rightarrow$Zh) & 5,560&567&1,466 \\
\hline
BMELD (Zh$\Rightarrow$En)  &4,427&517&1,135 \\
\bottomrule
\end{tabular}}
\vspace{5pt}
\begin{tablenotes}
	\item Train/Valid/Test splits corresponding to different usages and translation directions. WMT20 is for sentence-level pre-training on both En$\Leftrightarrow$De and En$\Leftrightarrow$Zh. Taskmaster-1 is the additional English dialogues, which is then translated to Germen and Chinese. BConTrasT and BMELD are used to fine-tune the NCT model on En$\Leftrightarrow$De and En$\Leftrightarrow$Zh, respectively.
\end{tablenotes}
\label{tab:dataset_statistics} 
\vspace{-10pt}
\end{table}

\subsection{Datasets}
As described in Section~\ref{subsec:model_training}, our proposed training framework consists of three stages, involving the large-scale sentence-level parallel corpus (WMT20), the additional monolingual dialogues (Taskmaster-1) and the annotated bilingual dialogues (BConTrasT and BMELD). Table~\ref{tab:dataset_statistics} lists the statistics of the involved datasets corresponding to different usages and translation directions.

\vspace{5pt}
\noindent\textbf{WMT20.\footnote{http://www.statmt.org/wmt20/translation-task.html}}
This large-scale sentence-level parallel corpus is used to at the first and second stages under our framework. For {English}$\Leftrightarrow${German}, we use and combine six corpora including Euporal, ParaCrawl, CommonCrawl, TildeRapid, NewsCommentary, and WikiMatrix. For En$\Leftrightarrow$Zh, the corpora we use contain News Commentary v15, Wiki Titles v2, UN Parallel Corpus V1.0, CCMT Corpus, and WikiMatrix. We first filter out duplicate sentence pairs and remove those whose length exceeds 80. Then, we employ a series of open-source/in-house scripts, including full-/half-width conversion, unicode conversion, punctuation normalization, and tokenization~\cite{DBLP:conf/wmt/WangTWDDS20} to pre-process the raw data. Finally, we apply byte-pair-encoding (BPE)~\cite{DBLP:conf/acl/SennrichHB16a} with 32K merge operations to tokenize the sentences into subwords. By doing so, we obtain 45,541,367 sentence pairs for En$\Leftrightarrow$De and 22,244,006 sentence pairs for En$\Leftrightarrow$Zh, respectively. 


\vspace{5pt}
\noindent\textbf{Taskmaster-1~\cite{DBLP:conf/emnlp/ByrneKSNGDYDKC19}.\footnote{https://github.com/google-research-datasets/Taskmaster/tree/master/TM-1-2019}}
The dataset~\cite{DBLP:conf/emnlp/ByrneKSNGDYDKC19} consists of English dialogues created via two distinct procedures, either the “Wizard of Oz” (WOz) approach in which trained agents and crowd-sourced workers interact with each other or the “self-dialog” where crowd-sourced workers write the entire dialog themselves. Given these monolingual dialogues in English, we first pre-process them using the same procedures as in WMT20. 
Then, because we do not have the needed German/Chinese monolingual dialogues in our En$\Leftrightarrow$De/En$\Leftrightarrow$Zh experiments, we use in-house En$\Rightarrow$De and En$\Rightarrow$Zh translation models to obtain the German/Chinese translations of those original English monolingual dialogues. 

\vspace{5pt}
\noindent\textbf{BConTrasT~\cite{DBLP:conf/wmt/FarajianLMMH20}.\footnote{https://github.com/Unbabel/BConTrasT}} This dataset is based on the monolingual Taskmaster-1 corpus~\cite{DBLP:conf/emnlp/ByrneKSNGDYDKC19} and is provided by WMT20 Shared Task on Chat Translation~\cite{DBLP:conf/wmt/FarajianLMMH20}, containing chats for the English-German language pair. A subset of dialogues in Taskmaster-1 are first automatically translated from English into German and then manually post-edited by native German speakers on Unbabel.\footnote{www.unbabel.com}
The conversations in BConTrasT involve two speakers of different languages, where one (customer) speaks in German and the other (agent) responds in English. 

\vspace{5pt}
\noindent\textbf{BMELD.} It is a recently released English-Chinese bilingual chat translation dataset. Based on the original English dialogues in MELD\footnote{The MELD is created by enhancing and extending EmotionLines dataset. It contains the same available dialogue instances in EmotionLines while encompassing audio and visual modality along with text.} (Multimodal EmotionLines Dataset)~\cite{DBLP:conf/acl/PoriaHMNCM19}, the dataset authors first crawl the corresponding Chinese translations from a movie subtitle website \footnote{https://www.zimutiantang.com/} and then manually post-edit these crawled translations by native post-graduate Chinese students majoring in English. Finally, following~\cite{DBLP:conf/wmt/FarajianLMMH20}, they assume 50\% of utterances are originally spoken by the Chinese speakers to keep data balance for Zh$\Rightarrow$En translations and build the \textbf{b}ilingual MELD (BMELD). For the Chinese utterances, we follow the authors to segment the sentences using Stanford CoreNLP toolkit.\footnote{https://stanfordnlp.github.io/CoreNLP/index.html}

\begin{table}[t]
\renewcommand\arraystretch{1.5}
\caption{Model Performance after Sentence-level Pre-training}
\centering
\setlength{\tabcolsep}{0.8mm}{
\begin{tabular}{l|c|c|c|c}
\toprule
\hline
\vspace{-2pt}
\textbf{Methods} & \textbf{En}$\Rightarrow$\textbf{De} & \textbf{De}$\Rightarrow$\textbf{En} & \textbf{En}$\Rightarrow$\textbf{Zh} & \textbf{Zh}$\Rightarrow$\textbf{En}
\\\hline
Transformer (Base)  &39.88  &40.72  &32.55  &24.42\\\hline
Transformer (Big)   &41.35  &41.56 &33.85  &24.86\\
\bottomrule
\end{tabular}}
\vspace{5pt}
\begin{tablenotes}
	\item The BLEU scores on \emph{newstest2019} of the model $M_{1}$ after sentence-level pre-training, corresponding to section~\ref{subsec:stage1}.
\end{tablenotes}
\label{tab:result_sent_nmt} 
\vspace{-10pt}
\end{table}

\subsection{Contrast Models}
\label{subsec:comp_models}
We compare the Flat-NCT model trained under our proposed MMT training framework with baseline sentence-level NMT models and several existing context-aware NMT models.
\vspace{10pt}

\noindent\textbf{Sentence-level NMT Models.}
\begin{itemize}
\setlength{\itemsep}{5pt}
\setlength{\parsep}{0pt}
\setlength{\parskip}{0pt}
\item \textbf{Transformer}~\cite{Vaswani:nips17}: The vanilla Transformer model trained on the sentence-level NMT corpus. 
\item \textbf{Transformer+FT}~\cite{Vaswani:nips17}: The vanilla Transformer model that is first pre-trained on the sentence-level NMT corpus and then directly fine-tuned on the bilingual chat translation dataset. 
\end{itemize}
\noindent\textbf{Context-Aware NMT Models.}
\begin{itemize}
\setlength{\itemsep}{5pt}
\setlength{\parsep}{0pt}
\setlength{\parskip}{0pt}
\item \textbf{Dia-Transformer+FT}~\cite{DBLP:conf/wmt/MarufMH18}: The original model is RNN-based document-level NMT model with an additional encoder to incorporate the mixed-language dialogue history. We re-implement it based on Transformer, where an additional encoder layer is used to incorporate the dialogue history into the NMT model.
\item \textbf{Gate-Transformer+FT}~\cite{DBLP:conf/emnlp/ZhangLSZXZL18}: A document-aware Transformer model that uses a gate to incorporate the context information.
\item \textbf{Flat-NCT+FT}: The Flat-NCT model trained through sentence-level pre-training (Section~\ref{subsec:sent_pretrain}) and context-aware fine-tuning (Section~\ref{subsec:context_ft}). Please note that it is our most related baseline.
\end{itemize}
\noindent\textbf{Our Model.}
\begin{itemize}
\item \textbf{Flat-NCT+MMT}: It is the Flat-NCT model trained under our proposed MMT training framework with Eq.~\ref{eq:stage3_new} used at the third stage, \emph{i.e.}, gradually transiting from monolingual to bilingual dialogues.
\end{itemize}

\subsection{Implementation Details}
We develop our NCT model based on the open-source toolkit THUMT.\footnote{https://github.com/THUNLP-MT/THUMT}~\cite{DBLP:conf/amta/TanZHCWSLL20} In experiments, we adopt the settings of~\emph{Transformer-Base} and \emph{Transformer-Big} as~\cite{Vaswani:nips17}. In \emph{Transformer-Base}, we use 512 as hidden size (\emph{i.e.}, $d$), 2,048 as filter size and 8 heads in multi-head attention. In \emph{Transformer-Big}, we use 1,024 as hidden size, 4,096 as filter size, and 16 heads in multi-head attention. Both \emph{Transformer-Base} and \emph{Transformer-Big} contain $L$$=$$6$ encoder layers and the identical number of decoder layers. 
As for the number of training steps for each stage, following the implementation in~\cite{liang-etal-2021-modeling}, we set the training steps of the first and second stages to 200,000 and 5,000, respectively. For the third stage, we conduct trial experiments on the En$\Rightarrow$De validation set, where the performance is no longer improved after about 5,000 steps. Therefore, we set the total training steps of the third training stage to 5,000, (\emph{i.e.}, $N$=5,000 in Eq.~\ref{eq:stage3_new}).

During training, we allocate 4,096 tokens to each NVIDIA Tesla V100 GPU. At the first stage, we use 8 GPUs to pre-train the model in parallel, resulting in 8*4,096 tokens per update. To test the performance of the pre-trained model, we measure its BLEU scores on \emph{newstest2019}. The results are shown in Table~\ref{tab:result_sent_nmt}. At the second and third stages, we only use 4 GPUs, resulting in about 4*4,096 tokens per update for all experiments at these two stages. All models are optimized using Adam~\cite{DBLP:journals/corr/KingmaB14} with the learning rate being 1.0 and label smoothing set to 0.1. The dropout rates for \emph{Transformer-Base} and \emph{Transformer-Big} are set to 0.1 and 0.3, respectively. The results are reported with the statistical significance test \cite{DBLP:conf/emnlp/Koehn04}.


\subsection{Effects of Hyper-paramerters}
For the Flat-NCT model under our proposed training framework, the context length for $\mathcal{C}_{\mathbf{x}_u}$ and the balancing factors ($\alpha_1$, $\beta_1$, $\alpha_2$ and $\beta_2$, see Eq.\ref{eq:stage2} and Eq.~\ref{eq:stage3_new}) of auxiliary tasks are the hyper-parameters we need to manually tune.

\begin{figure}[t]
\centering
\includegraphics[width=0.40\textwidth]{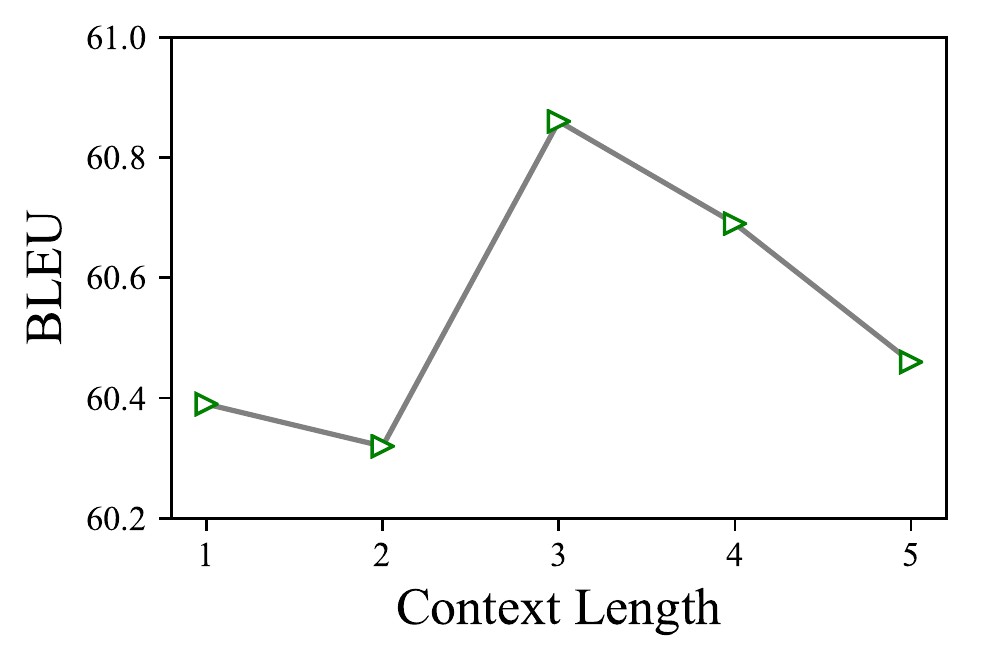}
\vspace{-10pt}
\caption{The effect of the context length for $\mathcal{C}_{\mathbf{x}_u}$. The BLEU scores of the Flat-NCT+FT model on the En$\Rightarrow$De validation set (under the \emph{Transformer-Base} setting).}
\label{fig:hyp_ctx_len}
\end{figure}

\subsubsection{Context Length}
In practice, for each $\textbf{x}_u$, the NCT model only takes a fixed length of preceding utterances as its dialogue history context $\mathcal{C}_{\textbf{x}_u}$. We investigate the effect of context length using the Flat-NCT+FT model with the \emph{Transformer-Base} setting. Fig.~\ref{fig:hyp_ctx_len} shows that the model achieves the best performance on the En$\Rightarrow$De validation set when the number of preceding source utterances for dialogue history context is set to 3. However, taking in more preceding utterances not only increases computational costs and but also adversely affects the performance. The underlying reason is that distant dialogue utterances usually have a low correlation with the current utterance and are likely to bring harmful noise.  Therefore, we set the context length to 3 in all subsequent experiments.

\begin{table}[]
\caption{Balancing Factor Determination}
	\centering
	\begin{tabular}{c|c|c|c|c}  
		\toprule
		\hline
		\vspace{-2pt}
		& & & & \\[-6pt]
		    & $\alpha_1$ & $\beta_1$ & $\alpha_2$ & $\beta_2$\\
		\hline
		& & & & \\[-6pt]
		\textbf{En}$\Rightarrow$\textbf{De} & 1.0 & 0.2 & 0.2 & 0.6 \\
		\hline
		& & & & \\[-6pt]
		\textbf{De}$\Rightarrow$\textbf{En} & 0.8 & 0.1 & 0.7 & 0.7 \\
		\hline& & & & \\[-6pt]
		\textbf{En}$\Rightarrow$\textbf{Zh} & 0.5 & 0.1 & 0.5 & 0.1 \\
		\hline& & & & \\[-6pt]
		\textbf{Zh}$\Rightarrow$\textbf{En} & 0.5 & 0.3 & 0.8 & 0.3 \\
		\bottomrule
	\end{tabular}
	\vspace{10pt}
\begin{tablenotes}
	\item The determined values of balancing factors for the auxiliary tasks.
\end{tablenotes}
\vspace{-5pt}
\label{tab:balancing_factors}
\end{table}

\subsubsection{Balancing Factors of Auxiliary Tasks}
To determine the best balancing factors ($\alpha_1$,$\beta_1$,$\alpha_2$,$\beta_2$) of auxiliary tasks, we evaluate the model performance on corresponding validation sets using the grid-search strategy. First, at the second training stage, we vary $\alpha_1$ and $\beta_1$ from 0 to 1.0 with the interval 0.1. Then, at the third training stage, given the selected $\alpha_1$ and $\beta_1$, we also search $\alpha_2$ and $\beta_2$ by drawing values from 0 to 1.0 with the interval 0.1. Finally, we obtain the sets of determined balancing factors for different translation directions (En$\Rightarrow$De, De$\Rightarrow$En, En$\Rightarrow$Zh and Zh$\Rightarrow$En), as listed in Table~\ref{tab:balancing_factors}. 

\begin{table*}[t!]
\renewcommand\arraystretch{1.5}
\caption{Overall Evaluation (BLEU$\uparrow$/TER$\downarrow$) of En$\Leftrightarrow$De and En$\Leftrightarrow$Zh Chat Translation Tasks}
\centering
\newcommand{\tabincell}[2]{\begin{tabular}{@{}#1@{}}#2\end{tabular}}
\setlength{\tabcolsep}{1.0mm}{
\begin{tabular}{c|l|ll|ll|ll|ll}
\toprule
\hline
\vspace{-2pt}
&\multicolumn{1}{c|}{\multirow{2}{*}{\textbf{Models (Base)}}} &\multicolumn{2}{c|}{\textbf{En}$\Rightarrow$\textbf{De}}  &  \multicolumn{2}{c|}{\textbf{De}$\Rightarrow$\textbf{En}}    &\multicolumn{2}{c|}{\textbf{En}$\Rightarrow$\textbf{Zh}}  &  \multicolumn{2}{c}{\textbf{Zh}$\Rightarrow$\textbf{En}} \\ 
&\multicolumn{1}{c|}{} & \multicolumn{1}{c}{BLEU$\uparrow$} & \multicolumn{1}{c|}{TER$\downarrow$} & \multicolumn{1}{c}{BLEU$\uparrow$} &  \multicolumn{1}{c|}{TER$\downarrow$}      & \multicolumn{1}{c}{BLEU$\uparrow$} & \multicolumn{1}{c|}{TER$\downarrow$} & \multicolumn{1}{c}{BLEU$\uparrow$} & \multicolumn{1}{c}{TER$\downarrow$}   \\ \hline
\multirow{2}{*}{\tabincell{c}{Sentence-level \\ NMT Models}}
&{Transformer} & 40.02 & 42.5 & 48.38 & 33.4 & 21.40 & 72.4 & 18.52 & 59.1  \\
&{Transformer+FT} & 58.43 & 26.7   & 59.57 & 26.2 & 25.22 & 62.8 & 21.59 & 56.7 \\\cdashline{1-10}[4pt/2pt]
\multirow{3}{*}{\tabincell{c}{Context-aware \\ NMT Models \quad }}
&{Dia-Transformer+FT} & 58.33 & 26.8 & 59.09 & 26.2 & 24.96 & 63.7 & 20.49 & 60.1   \\
&{Gate-Transformer+FT} & 58.48 & 26.6 & 59.53 & 26.1 & 25.34 & 62.5 & 21.03 & 56.9 \\
&{Flat-NCT+FT} & 58.15 & 27.1 & 59.46 & 25.7  & 24.76 & 63.4 & 20.61 & 59.8  \\
\cdashline{1-10}[4pt/2pt]
\multirow{1}{*}{\tabincell{c}{Our Model}}
&{Flat-NCT+MMT} & \textbf{59.33}$^{\dagger\dagger}$ & \textbf{26.2} & \textbf{60.17}$^{\dagger}$ & \textbf{25.1}$^{\dagger}$ & \textbf{27.43}$^{\dagger\dagger}$ & \textbf{60.4}$^{\dagger\dagger}$ & \textbf{22.21}$^{\dagger}$ & \textbf{56.1}$^{\dagger}$ \\ 
\hline
&\multicolumn{1}{c|}{\multirow{2}{*}{\textbf{Models (Big)}}} &\multicolumn{2}{c|}{\textbf{En}$\Rightarrow$\textbf{De}}  &  \multicolumn{2}{c|}{\textbf{De}$\Rightarrow$\textbf{En}}    &\multicolumn{2}{c|}{\textbf{En}$\Rightarrow$\textbf{Zh}}  &  \multicolumn{2}{c}{\textbf{Zh}$\Rightarrow$\textbf{En}} \\ 
&\multicolumn{1}{c|}{} & \multicolumn{1}{c}{BLEU$\uparrow$} & \multicolumn{1}{c|}{TER$\downarrow$} & \multicolumn{1}{c}{BLEU$\uparrow$} &  \multicolumn{1}{c|}{TER$\downarrow$}      & \multicolumn{1}{c}{BLEU$\uparrow$} & \multicolumn{1}{c|}{TER$\downarrow$} & \multicolumn{1}{c}{BLEU$\uparrow$} & \multicolumn{1}{c}{TER$\downarrow$}   \\ \hline
\multirow{2}{*}{\tabincell{c}{Sentence-level \\ NMT Models}}
&{Transformer} & 40.53 & 42.2 & 49.90 & 33.3 & 22.81 & 69.6 & 19.58 & 57.7 \\
&{Transformer+FT} & 59.01 & 26.0 & 59.98  & 25.9 & 26.95 & 60.7 & 22.15 & 56.1 \\\cdashline{1-10}[4pt/2pt]
\multirow{3}{*}{\tabincell{c}{Context-Aware\\ NMT Models}}
&{Dia-Transformer+FT} & 58.68 & 26.8  & 59.63 & 26.0 & 26.72 & 62.4 & 21.09 & 58.1  \\ 
&{Gate-Transformer+FT} & 58.94 & 26.2 & 60.08 & 25.5 & 27.10 & 60.3 & 22.26 & 55.8   \\
&{Flat-NCT+FT} & 58.61 & 26.5 & 59.98 & 25.4 & 26.45 & 62.6 & 21.38 & 57.7  \\
\cdashline{1-10}[4pt/2pt]
\multirow{1}{*}{\tabincell{c}{Our Model}}
&{Flat-NCT+MMT} & \textbf{60.11}$^{\dagger\dagger}$ & \textbf{25.8} & \textbf{61.04}$^{\dagger\dagger}$ & \textbf{25.0} & \textbf{28.62}$^{\dagger\dagger}$ & \textbf{59.6}$^{\dagger}$ & \textbf{23.08}$^{\dagger}$ & \textbf{54.9}$^{\dagger\dagger}$ \\
\bottomrule
\end{tabular}}
\vspace{10pt}
\begin{tablenotes}
	\item Results on the test sets of BConTrasT (En$\Leftrightarrow$De) and BMELD (En$\Leftrightarrow$Zh) in terms of BLEU (\%) and TER (\%). $\uparrow$: The higher the better. $\downarrow$: The lower the better. The best results are shown in bold. ``$^{\dagger}$'' and ``$^{\dagger\dagger}$'' indicate the results are statistically better than the best results of all other contrast NMT models with t-test {\em p} \textless \ 0.05 and {\em p} \textless \ 0.01, respectively. All the contrast models with "+FT" are trained using the conventional two-stage strategy. ``Flat-NCT+MMT'' is our model.
\end{tablenotes}
\vspace{-5pt}
\label{tab:main_result}
\end{table*}

\subsection{Overall Performance}
In Table~\ref{tab:main_result}, we report the experimental results on En$\Leftrightarrow$De and En$\Leftrightarrow$Zh using \emph{Transformer-Base} and \emph{Transformer-Big} settings. 

\subsubsection{Sentence-level Models v.s. Context-aware Models}
From Table~\ref{tab:main_result}, in terms of both BLEU and TER, we can observe that the sentence-level model ``Transformer+FT'' achieves comparable or even better results compared with those existing context-aware models (`Dia-Transformer+FT'', ``Gate-Transformer+FT'' and ``Flat-NCT+FT'') which are originally proposed for document-level translation. This suggests that if conventional approaches of exploiting context are not well adapted to the chat scenario, the NCT model would be negatively affected. This may be because when the size of training data for chat translation is extremely small, the NCT model is insufficiently trained and its poor use of dialogue history context adversely brings harmful noise.

\subsubsection{Results on En$\Leftrightarrow$De}
Under the \emph{Transformer-Base} setting, our NCT model outperforms sentence-level models and context-aware models in most cases. In terms of BLEU, compared with ``Flat-NCT+FT'', ``Flat-NCT+MMT'' performs 1.18$\uparrow$ on En$\Rightarrow$De and 0.71$\uparrow$ on De$\Rightarrow$En, showing the advantages of our proposed MMT training framework over the conventional two-stage training strategy. 
In terms of TER, ``Flat-NCT+MMT'' also exhibits its advantage over other contrast models. Under the \emph{Transformer-Big} setting, we can observe that ``Flat-NCT+MMT'' still performs the best in most cases on both En$\Rightarrow$De and De$\Rightarrow$En.

\subsubsection{Results on En$\Leftrightarrow$Zh}
We also conducted experiments on the BMELD dataset. Under the \emph{Transformer-Base} setting, on En$\Leftrightarrow$Zh, ``Flat-NCT+MMT'' substantially outperforms other sentence-level models and context-aware models. Concretely, ``Flat-NCT+MMT'' performs at least 2.09$\uparrow$ and 0.62$\uparrow$ BLEU scores over other contrast models on En$\Rightarrow$Zh and Zh$\Rightarrow$En, respectively. In terms of TER, it also achieves the best results in the two translation directions.
Under the \emph{Transformer-Big} setting, ``Flat-NCT+MMT'' exhibits notable performance gains again.

All the above results demonstrate the effectiveness and generalizability of our proposed MMT training framework across different language pairs.

\subsection{Result Analysis}
In order to better understand the advantages of our proposed training framework, we conduct a series of analytical experiments to investigate the effectiveness of using additional monolingual dialogues and the introduced auxiliary tasks.

\begin{table}[t]
\renewcommand\arraystretch{1.5}
\caption{Performance with Different Monolingual Dialogue Groups Removed}
\centering
\setlength{\tabcolsep}{0.6mm}{
\begin{tabular}{c|c|c|c|c|c}
\toprule
\hline
\vspace{-2pt}
&\multirow{2}{*}{\textbf{Models (Base)}} & 
\multicolumn{2}{c|}{\textbf{En}$\Rightarrow$\textbf{De}} & 
\multicolumn{2}{c}{\textbf{De}$\Rightarrow$\textbf{En}} \\
&&
\multicolumn{1}{c}{BLEU$\uparrow$} & 
\multicolumn{1}{c|}{TER$\downarrow$} & 
\multicolumn{1}{c}{BLEU$\uparrow$} &  
\multicolumn{1}{c}{TER$\downarrow$} \\
\hline
0 &\multicolumn{1}{c|}{Flat-NCT+MMT} &
\multicolumn{1}{c}{\textbf{60.86}} & \textbf{24.6} & \multicolumn{1}{c}{\textbf{60.94}} & \textbf{25.3} \\
\cdashline{1-6}
1 &\multicolumn{1}{l|}{\circled{3}: w\!/\!o. $\overline{X}$, $\overline{Y}$}  & \multicolumn{1}{c}{60.51} & \textbf{24.6} & \multicolumn{1}{c}{60.72} & 25.5 \\
2 &\multicolumn{1}{l|}{\circled{2}: w\!/\!o. $\overline{X}$, $\overline{Y}$} & \multicolumn{1}{c}{60.46} & 24.9 & \multicolumn{1}{c}{60.64} & 25.2 \\
3 &\multicolumn{1}{l|}{\circled{2}: w\!/\!o. $\overline{X}$ \,\,\,\,\,\,\, \circled{3}: w\!/\!o. $\overline{X}$} & \multicolumn{1}{c}{60.18}  & 24.9 & \multicolumn{1}{c}{60.50} & 25.8 \\
4 &\multicolumn{1}{l|}{\circled{2}: w\!/\!o. $\overline{Y}$ \,\,\,\quad \circled{3}: w\!/\!o. $\overline{Y}$} & \multicolumn{1}{c}{59.83}  & 25.3 & \multicolumn{1}{c}{59.69} & 25.9 \\
5 &\multicolumn{1}{l|}{\circled{2}: w\!/\!o. $\overline{X}$, $\overline{Y}$ \circled{3}: w\!/\!o. $\overline{X}$, $\overline{Y}$} & \multicolumn{1}{c}{59.74}  & 25.6 & \multicolumn{1}{c}{60.11} & 25.9 \\
\bottomrule
\end{tabular}}
\vspace{10pt}
\begin{tablenotes}
    \item Results on the validation set of BConTrasT (En$\Leftrightarrow$De) when different groups of monolingual dialogues are removed from MMT training framework. \circled{2}~and~\circled{3} denote the second and third training stages, respectively. ``w\!/\!o.'': the specified group of monolingual dialogues is removed. For instance, ``\circled{2}: w\!/\!o. $\overline{X}$, $\overline{Y}$'' means $\overline{X}$ and $\overline{Y}$ are removed at the second training stage.
\end{tablenotes}
\vspace{-5pt}
\label{tab:effect_md}
\end{table}

\begin{figure}[t]
\centering
\includegraphics[width=0.48\textwidth]{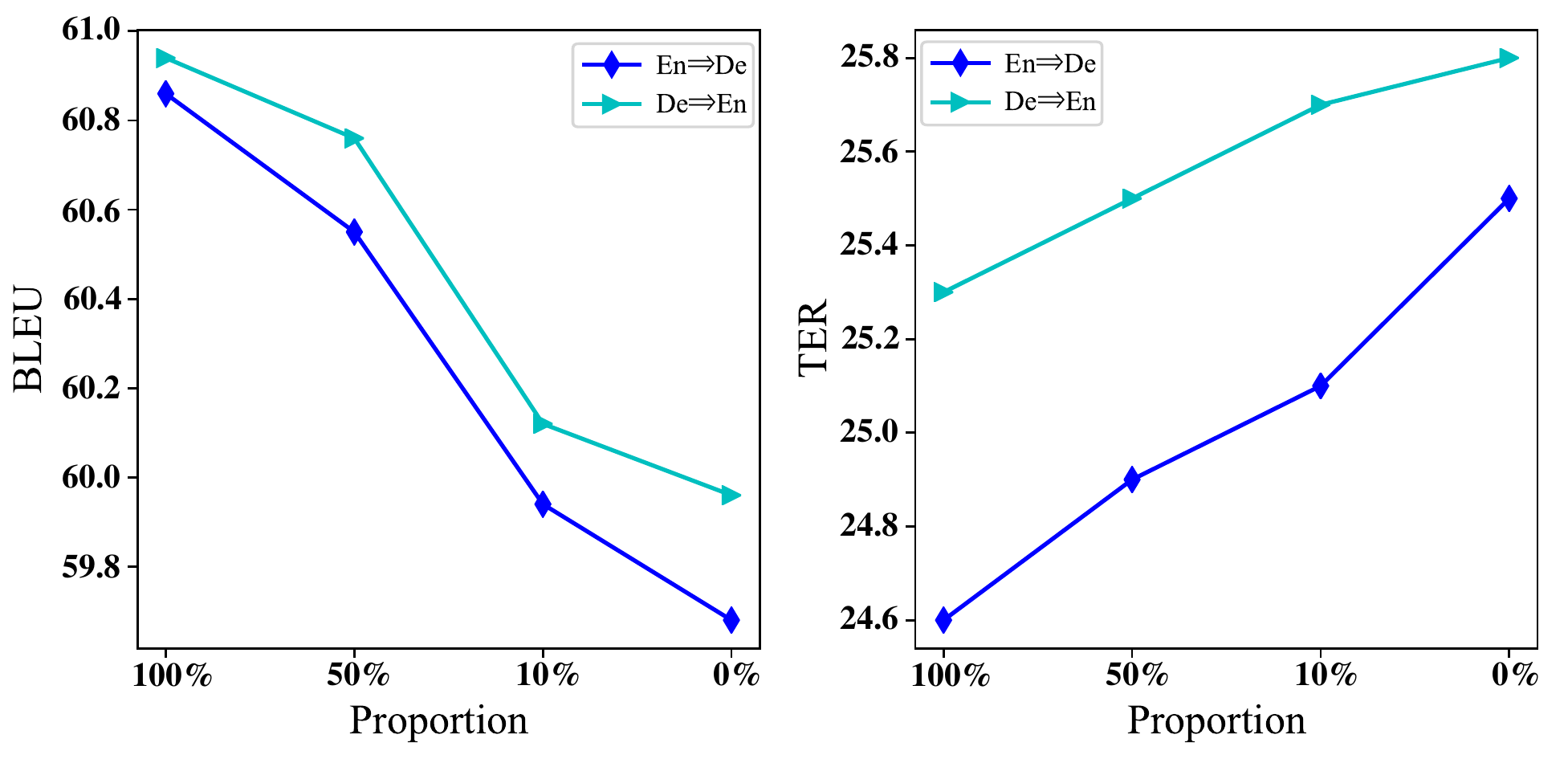}
\caption{Results (Left: BLEU$\uparrow$ / Right: TER$\downarrow$) on the validation set of BConTrasT (En$\Leftrightarrow$De) using different proportions of used monolingual dialogues (under the \emph{Transformer-Base} setting).}
\vspace{-10pt}
\label{fig:prop_md}
\end{figure}

\begin{table*}[t!]
\renewcommand\arraystretch{1.6}
\caption{Performance with Ablations of UD/SD Tasks}
\centering
\newcommand{\tabincell}[2]{\begin{tabular}{@{}#1@{}}#2\end{tabular}}
\setlength{\tabcolsep}{1.0mm}{
\begin{tabular}{c|l|ll|ll|l|ll|ll}
\toprule
\hline
\vspace{0pt}
& \multicolumn{5}{c|}{\textbf{UD}} & \multicolumn{5}{c}{\textbf{SD}} \\ 
\cline{2-11}
\vspace{-2pt}
& \multicolumn{1}{c|}{\multirow{2}{*}{\textbf{Models (Base)}}} &
\multicolumn{2}{c|}{\textbf{En}$\Rightarrow$\textbf{De}} &
\multicolumn{2}{c|}{\textbf{De}$\Rightarrow$\textbf{En}} &
\multicolumn{1}{c|}{\multirow{2}{*}{\textbf{Models (Base)}}} &
\multicolumn{2}{c|}{\textbf{En}$\Rightarrow$\textbf{De}} &  \multicolumn{2}{c}{\textbf{De}$\Rightarrow$\textbf{En}} \\ 
& & \multicolumn{1}{c}{BLEU$\uparrow$} & \multicolumn{1}{c|}{TER$\downarrow$} & \multicolumn{1}{c}{BLEU$\uparrow$} &  \multicolumn{1}{c|}{TER$\downarrow$} & & \multicolumn{1}{c}{BLEU$\uparrow$} & \multicolumn{1}{c|}{TER$\downarrow$} & \multicolumn{1}{c}{BLEU$\uparrow$} & \multicolumn{1}{c}{TER$\downarrow$}   \\ \hline
\multirow{1}{*}{\tabincell{c}{0}}
&{Flat-NCT+MMT} & \textbf{60.86} & \textbf{24.6} & \textbf{60.94} & \textbf{25.3} & {Flat-NCT+MMT} & \textbf{60.86} & \textbf{24.6} & \textbf{60.94} & \textbf{25.3}  \\ 
\cdashline{1-11}[4pt/2pt]
\multirow{1}{*}{\tabincell{c}{1}}
&{w\!/\!o. $\mathcal{L}_{ud}^{\overline{X}}$} & 60.80 & 24.7 & 60.72 & 25.7 & {w\!/\!o. $\mathcal{L}_{sd}^{\overline{X}}$} & 60.51 & 25.0 & 60.43 & 26.1  \\
\multirow{1}{*}{\tabincell{c}{2}}
&{w\!/\!o. $\mathcal{L}_{ud}^{\overline{Y}}$} & 60.47 & 24.9 & 60.43 & 26.1 & {w\!/\!o. $\mathcal{L}_{sd}^{\overline{Y}}$} & 60.29 & 24.7 & 60.83 & 25.6  \\
\multirow{1}{*}{\tabincell{c}{3}}
&{w\!/\!o. $\mathcal{L}_{ud}^{\overline{X}}$,$\mathcal{L}_{ud}^{\overline{Y}}$} & 59.96 & 25.3 & 60.41 & 25.9 & {w\!/\!o. $\mathcal{L}_{sd}^{\overline{X}}$,$\mathcal{L}_{sd}^{\overline{Y}}$} & 60.13 & 25.0 & 60.66 & 25.6  \\
\multirow{1}{*}{\tabincell{c}{4}}
&{w\!/\!o. $\mathcal{L}_{ud}^{X}$} & 60.43 & 24.9 & 60.20 & 26.1 & {w\!/\!o. $\mathcal{L}_{sd}^{X}$} & 60.36 & 25.2 & 60.76 & 26.0  \\
\multirow{1}{*}{\tabincell{c}{5}}
&{w\!/\!o. $\mathcal{L}_{ud}^{Y}$} & 60.25 & 24.8 & 60.56 & 25.5 & {w\!/\!o. $\mathcal{L}_{sd}^{Y}$} & 60.22 & 25.0 & 60.47 & 26.0 \\
\multirow{1}{*}{\tabincell{c}{6}}
&{w\!/\!o. $\mathcal{L}_{ud}^{X}$,$\mathcal{L}_{ud}^{Y}$} & 59.89 & 25.1 & 60.25 & 25.7 & {w\!/\!o. $\mathcal{L}_{sd}^{X}$,$\mathcal{L}_{sd}^{Y}$} & 60.27 & 25.3 & 60.56 & 25.3 \\
\multirow{1}{*}{\tabincell{c}{7}}
&{w\!/\!o. $\mathcal{L}_{ud}^{\overline{X}}$,$\mathcal{L}_{ud}^{\overline{Y}}$,$\mathcal{L}_{ud}^{X}$,$\mathcal{L}_{ud}^{Y}$} & 59.86 & 25.3 & 60.04 & 26.0 & {w\!/\!o. $\mathcal{L}_{sd}^{\overline{X}}$,$\mathcal{L}_{sd}^{\overline{Y}}$,$\mathcal{L}_{sd}^{X}$,$\mathcal{L}_{sd}^{Y}$} & 59.97 & 25.5 & 60.39 & 25.9  \\
\multirow{1}{*}{\tabincell{c}{8}}
&{w\!/\!o. any UD/SD task} & 59.79 & 25.5 & 59.97 & 26.5 & {w\!/\!o. any UD/SD task} & 59.79 & 25.5 & 59.97 & 26.5  \\

\bottomrule
\end{tabular}}
\vspace{10pt}
\begin{tablenotes}
	\item Results (BLEU$\uparrow$/TER$\downarrow$) on the validation set of BConTrasT (En$\Leftrightarrow$De) with ablations of UD/SD tasks. The left half lists ablation results of the UD task while the right lists those of the SD task. ``w\!/\!o.'': the specified training objectives are ablated in our proposed training framework. For instance, ``w\!/\!o. $\mathcal{L}_{ud}^{\overline{X}}$'' means the objective of the UD task $\mathcal{L}_{ud}^{\overline{X}}$ on source-language monolingual dialogues $\overline{X}$ is ablated in Eq.~\ref{eq:stage2} and Eq.~\ref{eq:stage3_new} at the second and third training stages. The last row (Row 8) corresponds to the setting that all the training objectives of auxiliary tasks are ablated, \emph{i.e.}, w\!/\!o. $\mathcal{L}_{ud}^{\overline{X}}$, $\mathcal{L}_{ud}^{\overline{Y}}$, $\mathcal{L}_{ud}^{X}$,
	$\mathcal{L}_{ud}^{Y}$,
	$\mathcal{L}_{sd}^{\overline{X}}$, $\mathcal{L}_{sd}^{\overline{Y}}$, $\mathcal{L}_{sd}^{X}$,
	$\mathcal{L}_{sd}^{Y}$.
\end{tablenotes}
\vspace{0pt}
\label{tab:ablation_UD_SD}
\end{table*}

\subsubsection{Effects of Monolingual Dialogues}
In our proposed training framework, we use both source- and target-language additional monolingual dialogues ($\overline{X}$ and $\overline{Y}$) at the second and third stages. 

First, we investigate the effect of monolingual dialogues on En$\Leftrightarrow$De validation set by partially removing different groups of them. From Table~\ref{tab:effect_md}, according to training stages, we can observe that the removal of monolingual dialogues at either the second or the third stage results in performance drops (Rows 1 and 2). This indicates that the additional monolingual dialogues benefit the NCT model at both training stages. Next, according to languages, when we totally remove one of the source-language and target-language monolingual dialogues at the two stages, the model performance also declines (Rows 3 and 4). These two results show that both the source- and target-language monolingual dialogues take positive effects during training. Lastly, if there is no monolingual data used during the whole training process, the performance degrades more drastically (Row 5), echoing those aforementioned findings again.

Then, we investigate how the amount of additional monolingual dialogues affects the NCT model. Fig.~\ref{fig:prop_md} illustrates the model performance with different proportions (100\%, 50\%, 10\% and 0\%) of used monolingual dialogues. The results show that the performance of the NCT model consistently declines with fewer monolingual dialogues used in our proposed training framework. All these results demonstrate the effectiveness and necessity  of using relatively abundant monolingual dialogues in our framework.


\subsubsection{Effects of Auxiliary Tasks}
The two auxiliary tasks (UD and SD) play an important role in our proposed training framework. Therefore, we investigate their effects by ablating them with different settings. Table~\ref{tab:ablation_UD_SD} lists the results on the validation set of BConTrasT (En$\Leftrightarrow$De) with ablations of UD/SD tasks. 

First, we successively exclude the objectives of UD/SD task on monolingual dialogues from the MMT training of our NCT model. When only one of $\mathcal{L}_{ud}^{\overline{X}}$, $\mathcal{L}_{ud}^{\overline{Y}}$, $\mathcal{L}_{sd}^{\overline{X}}$ and $\mathcal{L}_{sd}^{\overline{Y}}$ is excluded, the performance drops (Rows 1 and 2) compared to ``Flat-NCT+MMT'' (Row 0). Moreover, if we exclude the UD or SD task on both source- and target-language monolingual dialogues at a time, the NCT model mostly performs worse than the above results (\emph{i.e.}, Row 3 v.s. Rows 0,1,2). It is also notable that the ablations of UD/SD tasks have a greater influence on En$\Rightarrow$De direction than on De$\Rightarrow$En. We conjecture that German monolingual dialogues are manually translated from English by in-house sentence-level NMT models, losing their original conversational properties to some extent. Thus, the two dialogue-related auxiliary tasks bring smaller improvements in the process of MMT training. These results show both UD and SD tasks on source- and target-language monolingual dialogues bring improvements, indicating that the preliminary capability of capturing dialogue context acquired from additional monolingual dialogues actually enhances the NCT model. 

Then, we turn to successively exclude the objective of UD/SD task on bilingual dialogues. We can obtain the similar conclusion that the exclusions of $\mathcal{L}_{ud}^{X}$ and $\mathcal{L}_{ud}^{Y}$ lead to the performance decline (\emph{i.e.}, Row 0 v.s. Rows 4,5,6). Similarly, the two auxiliary tasks on source- and target-language bilingual dialogues take greater effects in most cases on En$\Rightarrow$De direction than on De$\Rightarrow$En, supporting the above-mentioned conjecture again.

Lastly, we completely ablate either the UD or SD task from the MMT training. We can observe that the performance drops more severely (Row 7). Moreover, if we totally remove all auxiliary objectives of UD and SD tasks, the training of our NCT model degenerates into the conventional two-stage training, thus obtaining the worst performance (Row 8). These ablation results with different settings strongly confirm that the two auxiliary tasks take considerable effects during the MMT training by incorporating the modelling of conversational properties into our NCT model.

\begin{table}[t]
\renewcommand\arraystretch{1.5}
\caption{Performance with Pseudo/Authentic Monolingual Dialogues}
\centering
\newcommand{\tabincell}[2]{\begin{tabular}{@{}#1@{}}#2\end{tabular}}
\setlength{\tabcolsep}{0.8mm}{
\begin{tabular}{c|ll|ll}
\toprule
\hline
\vspace{-2pt}
{\multirow{2}{*}{\textbf{Models (Base)}}} &\multicolumn{2}{c|}{\textbf{En}$\Rightarrow$\textbf{Zh}}  &\multicolumn{2}{c}{\textbf{Zh}$\Rightarrow$\textbf{En}} \\ 
\multicolumn{1}{c|}{} & \multicolumn{1}{l}{BLEU$\uparrow$} & \multicolumn{1}{l|}{TER$\downarrow$} & \multicolumn{1}{l}{BLEU$\uparrow$} & \multicolumn{1}{l}{TER$\downarrow$}   \\ \hline
\multicolumn{1}{l|}{Flat-NCT+MMT(Pseudo) w\!/\!o. SD} & 27.35 & 60.6 & 22.12 & 56.4  \\
\multicolumn{1}{l|}{Flat-NCT+MMT(Authentic) w\!/\!o. SD} & 27.80 & 59.7 & 22.82 & 55.8 \\
\hline
\vspace{0pt}
{\multirow{2}{*}{\textbf{Models (Big)}}} &\multicolumn{2}{c|}{\textbf{En}$\Rightarrow$\textbf{Zh}}  &\multicolumn{2}{c}{\textbf{Zh}$\Rightarrow$\textbf{En}} \\ 
\multicolumn{1}{c|}{} & \multicolumn{1}{l}{BLEU$\uparrow$} & \multicolumn{1}{l|}{TER$\downarrow$} & \multicolumn{1}{l}{BLEU$\uparrow$} & \multicolumn{1}{l}{TER$\downarrow$}   \\ \hline
\multicolumn{1}{l|}{Flat-NCT+MMT(Pseudo) w\!/\!o. SD} & 28.31 & 59.7 & 22.87 & 55.3  \\
\multicolumn{1}{l|}{Flat-NCT+MMT(Authentic) w\!/\!o. SD} & 28.55 & 59.0 & 23.36 & 54.0 \\
\bottomrule
\end{tabular}}
\vspace{10pt}
\begin{tablenotes}
	\item Results on the test set of BMELD (En$\Leftrightarrow$Zh) in terms of BLEU (\%) and TER (\%). ``Flat-NCT+MMT(Pseudo) w\!/\!o. SD'' represents using pseudo Chinese monolingual dialogues without any SD objective. ``Flat-NCT+MMT(Authentic) w\!/\!o. SD'' represents using authentic Chinese monolingual dialogues without any SD objective.
\end{tablenotes}
\vspace{-5pt}
\label{tab:effect_authentic}
\end{table}

\subsubsection{Effects of Pseudo/Authentic Monolingual Dialogues}
In our previous experiments, since most German and Chinese dialogue datasets do not contain annotated speaker labels, they are not suitable for our Flat-NCT model to accomplish SD task. Therefore, we use in-house NMT models to obtain pseudo German/Chinese monolingual dialogues from authentic English Taskmaster-1 dataset that has available speaker labels. To investigate how the authenticity of monolingual dialogues would affect our proposed training framework, we turn to use totally authentic monolingual dialogues.

Specifically, besides the authentic English Taskmaster-1 dataset, we introduce the authentic Chinese dialogues from the recently-released MSCTD dataset~\cite{liang-etal-2022-msctd}.\footnote{MSCTD dataset has a total of 132,741 Chinese utterances.} When using MSCTD dataset, as it still has no speaker label for SD task, we only include the UD task, \emph{i.e.}, excluding $\mathcal{L}_{sd}^{\overline{X}}$, $\mathcal{L}_{sd}^{\overline{Y}}$, $\mathcal{L}_{sd}^{X}$, $\mathcal{L}_{sd}^{Y}$ from MMT training, which is denoted as ``Flat-NCT+MMT(Authentic) w\!/\!o. SD''. Table~\ref{tab:effect_authentic} gives its comparison with the model using pseudo Chinese monolingual dialogues, \emph{i.e.}, ``Flat-NCT+MMT(Pseudo) w\!/\!o. SD''. From the table, we can see that ``Flat-NCT+MMT(Authentic) w\!/\!o. SD'' outperforms ``Flat-NCT+MMT(Pseudo) w\!/\!o. SD'' under both the \emph{Transformer-Base} and \emph{Transformer-Big} settings. This shows authentic monolingual dialogues are indeed more beneficial to the NCT model, indicating that our MMT training framework has the potential to further boost model performance if there are suitable monolingual dialogue datasets with speaker labels on both source and target languages.

\begin{table}[t]
\renewcommand\arraystretch{1.5}
\caption{Performance with BT-augmented Chat Translation Corpus $D'_{bct}$}
\centering
\newcommand{\tabincell}[2]{\begin{tabular}{@{}#1@{}}#2\end{tabular}}
\setlength{\tabcolsep}{0.8mm}{
\begin{tabular}{c|ll|ll}
\toprule
\hline
\vspace{-2pt}
{\multirow{2}{*}{\textbf{Models (Base)}}} &\multicolumn{2}{c|}{\textbf{En}$\Rightarrow$\textbf{Zh}}  &\multicolumn{2}{c}{\textbf{Zh}$\Rightarrow$\textbf{En}} \\ 
\multicolumn{1}{c|}{} & \multicolumn{1}{l}{BLEU$\uparrow$} & \multicolumn{1}{l|}{TER$\downarrow$} & \multicolumn{1}{l}{BLEU$\uparrow$} & \multicolumn{1}{l}{TER$\downarrow$}   \\ \hline
\multicolumn{1}{l|}{Transformer + FT($D'_{bct}$)} & 26.04 & 61.7 & 21.77 & 56.2  \\
\multicolumn{1}{l|}{Gate-Transformer + FT($D'_{bct}$)} & 26.36 & 61.2 & 21.61 & 55.8  \\
\multicolumn{1}{l|}{Flat-NCT+MMT($D'_{bct}$) w\!/\!o. SD} & 28.15 & 59.6 & 22.44 & 55.6 \\
\hline
\vspace{0pt}
{\multirow{2}{*}{\textbf{Models (Big)}}} &\multicolumn{2}{c|}{\textbf{En}$\Rightarrow$\textbf{Zh}}  &\multicolumn{2}{c}{\textbf{Zh}$\Rightarrow$\textbf{En}} \\ 
\multicolumn{1}{c|}{} & \multicolumn{1}{l}{BLEU$\uparrow$} & \multicolumn{1}{l|}{TER$\downarrow$} & \multicolumn{1}{l}{BLEU$\uparrow$} & \multicolumn{1}{l}{TER$\downarrow$}   \\ \hline
\multicolumn{1}{l|}{Transformer + FT($D'_{bct}$)} & 27.29 & 60.3 & 22.38 & 55.9  \\
\multicolumn{1}{l|}{Gate-Transformer + FT($D'_{bct}$)} & 27.65 & 59.9 & 22.45 & 55.6  \\
\multicolumn{1}{l|}{Flat-NCT+MMT($D'_{bct}$) w\!/\!o. SD} & 28.81 & 58.7 & 23.17 & 55.1 \\
\bottomrule
\end{tabular}}
\vspace{10pt}
\begin{tablenotes}
	\item Results on the test set of BMELD (En$\Leftrightarrow$Zh) in terms of BLEU (\%) and TER (\%). ``Transformer + FT($D'_{bct}$)'' and ``Gate-Transformer + FT($D'_{bct}$)'' represents using the BT-augmented dataset $D'_{bct}$ to fine-tune the Transformer model and Gate-Transformer model ,respectively. ``Flat-NCT+MMT($D'_{bct}$) w\!/\!o. SD'' represents using $D'_{bct}$ to train the Flat-NCT model through MMT training framework without any SD objective.
\end{tablenotes}
\vspace{-10pt}
\label{tab:effect_bt_augment}
\end{table}

\subsubsection{Effects of BT-augmented Chat Translation Corpus}
Instead of just being used for the auxiliary tasks, the additional monolingual dialogues can be alternatively used to augment the bilingual chat translation dataset $D_{bct}$ for the context-aware fine-tuning of all contrast models and ours. To further validate the effectiveness of our proposed training framework, we make comparisons between MMT training and conventional two-stage pretrain-finetune paradigm using BT-augmented bilingual chat translation dataset.

Concretely, as a common technique, we employ back-translation to augment the original dataset $D_{bct}$ to $D'_{bct}$. For En$\Rightarrow$Zh, the target-side additional Chinese dialogues from MSCTD dataset are translated into English. Conversely, for Zh$\Rightarrow$En, the target-side English additional dialogues from Taskmaster-1 dataset are translated into Chinese. Due to the lack of speaker labels in MSCTD dataset, we also exclude all SD objectives in MMT training and compare ``Flat-NCT+MMT($D'_{bct}$) w\!/\!o. SD'' with the sentence-level ``Transformer+FT($D'_{bct}$)'' and ``Gate-Transformer+FT($D'_{bct}$)''.~\footnote{``Gate-Transformer + FT'' is chosen because it is the most competitive among all context-aware contrast models with two-stage training, as shown in Table~\ref{tab:main_result}.} From Table~\ref{tab:effect_bt_augment}, we can observe ``Flat-NCT+MMT($D'_{bct}$) w\!/\!o. SD'' outperforms ``Transformer+FT($D'_{bct}$)'' and ``Gate-Transformer + FT($D'_{bct}$)'' under both \emph{Transformer-Base} and \emph{Transformer-Big} settings, which demonstrates that our proposed training framework can still take notable effects when the bilingual chat translation corpus for context-aware fine-tuning is adequately augmented.

\begin{table}[t]
\renewcommand\arraystretch{1.5}
\caption{Performance with/without Gradual Transition Strategy}
\centering
\newcommand{\tabincell}[2]{\begin{tabular}{@{}#1@{}}#2\end{tabular}}
\setlength{\tabcolsep}{0.8mm}{
\begin{tabular}{c|ll|ll}
\toprule
\hline
\vspace{-2pt}
{\multirow{2}{*}{\textbf{Models (Big)}}} &\multicolumn{2}{c|}{\textbf{En}$\Rightarrow$\textbf{De}}  &\multicolumn{2}{c}{\textbf{De}$\Rightarrow$\textbf{En}} \\ 
\multicolumn{1}{c|}{} & \multicolumn{1}{l}{BLEU$\uparrow$} & \multicolumn{1}{l|}{TER$\downarrow$} & \multicolumn{1}{l}{BLEU$\uparrow$} & \multicolumn{1}{l}{TER$\downarrow$}   \\ \hline
\multicolumn{1}{l|}{Flat-NCT+MMT} & 60.11 & 25.8 & 61.04 & 25.0  \\
\multicolumn{1}{l|}{Flat-NCT+MMT w\!/\!o. GT} & 59.62 & 26.2 & 60.76 & 25.2 \\
\hline
{\multirow{2}{*}{\textbf{Models (Big)}}} &\multicolumn{2}{c|}{\textbf{En}$\Rightarrow$\textbf{Zh}}  &\multicolumn{2}{c}{\textbf{Zh}$\Rightarrow$\textbf{En}} \\ 
\multicolumn{1}{c|}{} & \multicolumn{1}{l}{BLEU$\uparrow$} & \multicolumn{1}{l|}{TER$\downarrow$} & \multicolumn{1}{l}{BLEU$\uparrow$} & \multicolumn{1}{l}{TER$\downarrow$}   \\ \hline
\multicolumn{1}{l|}{Flat-NCT+MMT} & 28.62 & 59.6 & 23.08 & 54.9  \\
\multicolumn{1}{l|}{Flat-NCT+MMT w\!/\!o. GT} & 28.18 & 59.8 & 22.50 & 55.9 \\
\bottomrule
\end{tabular}}
\vspace{10pt}
\begin{tablenotes}
	\item Results on the test sets of BConTrasT (En$\Leftrightarrow$De) and BMELD (En$\Leftrightarrow$Zh) in terms of BLEU (\%) and TER (\%). ``Flat-NCT+MMT'': the Flat-NCT model trained using the gradual transition strategy from monolingual to bilingual dialogues (Eq.~\ref{eq:stage3_new}). ``Flat-NCT+MMT w\!/\!o. GT'': the Flat-NCT model trained without using the gradual transition strategy (Eq.~\ref{eq:stage3}).
\end{tablenotes}
\vspace{-5pt}
\label{tab:effect_GT}
\end{table}

\subsubsection{Effects of Gradual Transition Strategy}
At the third stage of our proposed framework, the Flat-NCT model is trained through Eq.~\ref{eq:stage3_new}, \emph{i.e.}, gradually transiting from using monolingual to bilingual dialogues. This strategy makes the transition from the second to the third stage smoother, which further alleviates the training discrepancy described in Section~\ref{subsubsec:stage3}. 

To investigate its effectiveness, we also train the NCT model through Eq.~\ref{eq:stage3}, \emph{i.e.}, without the strategy of gradual transition. As shown in Table~\ref{tab:effect_GT}, under the \emph{Transformer-Big} setting, the performance of ``Flat-NCT+MMT w\!/\!o. GT'' is significantly worse than those of ``Flat-NCT+MMT'' across all translation directions. These results indicate that the gradual transition strategy makes better use of additional monolingual dialogues, benefiting the training of our NCT model.

\subsection{Evaluation of Translation Quality}
To further verify the benefits of our proposed training framework, we assess the quality of translations generated by different NCT models using automatic and human evaluations.

\subsubsection{Automatic Evaluation of Dialogue Coherence}
Following \cite{DBLP:conf/ijcai/LapataB05,DBLP:conf/aaai/XiongH0W19}, we use the cosine similarity between each translated utterance $\mathbf{x}_u$ and its corresponding dialogue context $\mathcal{C}_{\mathbf{x}_u}$ to automatically measure dialogue coherence, which is defined as 
\begin{equation}\nonumber
\setlength{\abovedisplayskip}{10pt}
\setlength{\belowdisplayskip}{10pt}
\begin{split}
    sim(\mathbf{x}_u, \mathcal{C}_{\mathbf{x}_u}) &= \mathrm{cos\_sim}(f(\mathbf{x}_u), f(\mathcal{C}_{\mathbf{x}_u})),
\end{split}
\end{equation}
where $f(\cdot)$ denotes the sequence representation obtained by averaging the word vectors of its included tokens. We use Word2Vec\footnote{https://code.google.com/archive/p/word2vec/}~\cite{DBLP:journals/corr/abs-1301-3781} trained on Taskmaster-1\footnote{The English utterances in BConTrasT comes from Taskmaster-1.} to obtain the distributed word vectors whose dimension is set to 100.

Table~\ref{tab:auto_coherence} shows the measured coherence of translated utterances with their corresponding dialogue context on the De$\Rightarrow$En test set of BConTrasT. It shows that our ``Flat-NCT+MMT'' produces more coherent translations compared to other contrast models (significance test, {\em p} \textless \ 0.01).

\begin{table}[t]
\renewcommand\arraystretch{1.5}
\caption{Automatic Evaluation of Dialogue Coherence}
\centering
\newcommand{\tabincell}[2]{
\begin{tabular}{@{}#1@{}}#2\end{tabular}}
\setlength{\tabcolsep}{1.0mm}{
\begin{tabular}{l|c|c|c|c}
\toprule
\hline
\vspace{-2pt}
\multirow{1}{*}{\textbf{Models (Base)}} &  \multicolumn{1}{c|}{\textbf{1-th Pr.}} & \multicolumn{1}{c|}{\textbf{2-th Pr.}}  & \multicolumn{1}{c|}{\textbf{3-th Pr.}}& \multicolumn{1}{c}{\textbf{ctx.}}\\\cline{1-5}
Transformer               &0.650  &0.604 &0.566 &0.612\\
Transformer+FT            &0.658  &0.610 &0.571 &0.619 \\\cdashline{1-5}[4pt/2pt]
Dia-Transformer+FT        &0.655  &0.608 &0.571 &0.617\\
{Gate-Transformer+FT}     &0.660  &0.614 &0.575 &0.620\\
Flat-NCT+FT               &0.657  &0.610 &0.571 &0.616\\
\cdashline{1-5}[4pt/2pt]
Flat-NCT+MMT &0.665$^{\dagger\dagger}$  &{0.617}$^{\dagger\dagger}$ &{0.578}$^{\dagger\dagger}$ &{0.629}$^{\dagger\dagger}$\\

\cdashline{1-5}[4pt/2pt]
Human Reference         &\textbf{0.666}  &\textbf{0.620} &\textbf{0.580} &\textbf{0.633}\\
\bottomrule
\end{tabular}}
\vspace{10pt}
\begin{tablenotes}
	\item Results of dialogue coherence in terms of sentence similarity (-1$\sim$1) on the test set of BConTrasT in De$\Rightarrow$En direction under the \emph{Transformer-Base} setting. The ``\#\textbf{-th Pr.}'' denotes the \#-th preceding utterance to the current one. ``$^{\dagger\dagger}$'' indicates the improvement over the best result of all other contrast models is statistically significant (\emph{p} \textless \ 0.01).
\end{tablenotes}
\label{tab:auto_coherence}
\vspace{-10pt}
\end{table}

\begin{table}[t]
\renewcommand\arraystretch{1.5}
\centering
\setlength{\tabcolsep}{1.8mm}{
\caption{Human Evaluation}
\begin{tabular}{l|c|c|c}
\toprule
\hline
\vspace{-2pt}
\multirow{1}{*}{\textbf{Models (Base)}} & \multicolumn{1}{c|}{$\textbf{DC.}$} &  \multicolumn{1}{c|}{$\textbf{SC.}$} &  \multicolumn{1}{c}{$\textbf{Flu.}$} \\\cline{1-4}
Transformer        &0.540 &0.485  &0.590 \\
Transformer+FT     &0.590 &0.530  &0.635 \\\cdashline{1-4}[4pt/2pt]
Dia-Transformer+FT &0.580   &0.525  &0.625 \\
{Gate-Transformer+FT}  &0.605     &0.540   &0.635 \\
Flat-NCT+FT &0.595   &0.525  &0.630 \\\cdashline{1-4}[4pt/2pt]
Flat-NCT+MMT       &\textbf{0.640} &\textbf{0.570}  &\textbf{0.665} \\
\bottomrule
\end{tabular}
\label{tab:human_evaluation}}
\vspace{10pt}
\begin{tablenotes}
	\item Results on the test set of BMELD ({Zh$\Rightarrow$En}) under the \emph{Transformer-Base} setting. ``\textbf{DC.}'': Dialogue Coherence. ``\textbf{SC.}'': Speaker Consistency. ``\textbf{Flu.}'': Fluency. The values for these three criteria range from 0 to 1.
\end{tablenotes}
\vspace{-10pt}
\end{table}

\subsubsection{Human Evaluation} Table~\ref{tab:human_evaluation} lists the results of human evaluation on the test set of BMELD (Zh$\Rightarrow$En). Following \cite{DBLP:conf/wmt/BaoSSLC20,DBLP:conf/wmt/FarajianLMMH20}, we conduct evaluations using three criteria: 1)~\textbf{Dialogue Coherence} (\textbf{DC.}) measures whether the translation is semantically coherent with the dialogue history context in a chat; 2)~\textbf{Speaker Consistency} (\textbf{SC.}) evaluates whether the translation preserves the characteristic of its original speaker; 3)~\textbf{Fluency} (\textbf{Flu.}) measures whether the translation is fluent and grammatically correct.

First, we randomly sample 200 dialogues from the test set of BMELD in Zh$\Rightarrow$En direction. Then, we use each of the models in Table~\ref{tab:human_evaluation} to generate the translations of these sampled dialogues. Finally, we assign these translated utterances and their corresponding dialogues in the target language to three postgraduate evaluators who are native Chinese speakers majoring in English with qualified certificates, and ask them to assess the translations according to the above three criteria.

The results in Table~\ref{tab:human_evaluation} show that the generated translation of our model (``Flat-NCT+MMT'') is more coherent to corresponding dialogue context, better preserves the characteristic of original speakers and is more fluent as well, indicating the superiority of our model. The inter-annotator agreements calculated by the Fleiss’ kappa~\cite{doi:10.1177/001316447303300309} are 0.535, 0.507, and 0.548 for {DC.}, {SC.} and {Flu.}, respectively.

\begin{figure*}[ht]
\centering
\includegraphics[width=0.96\textwidth]{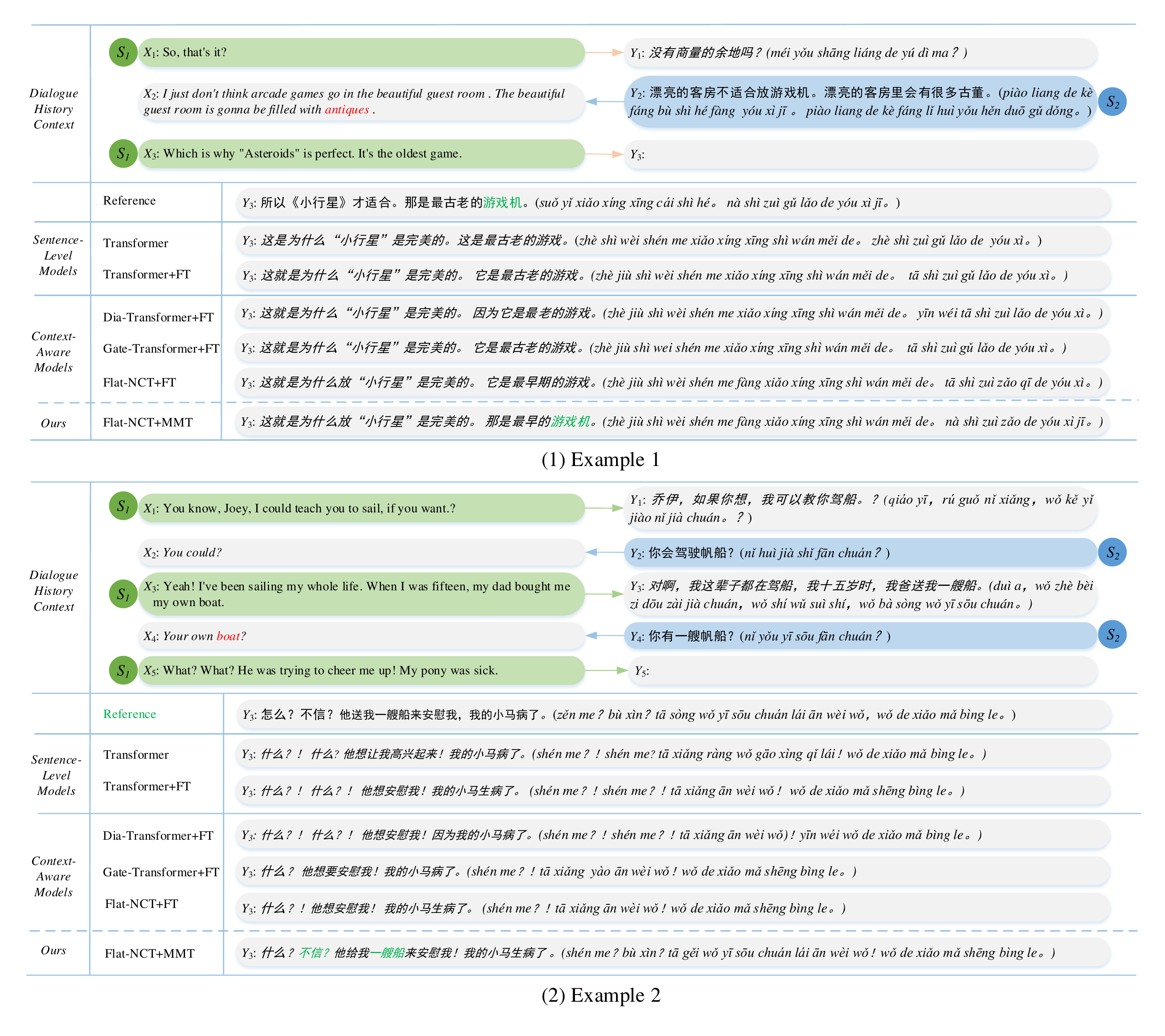}
\caption{Two illustrative case examples from the test set of BMELD (En$\Rightarrow$Zh).}
\label{fig:case_study}
\vspace{-6pt}
\end{figure*}

\subsubsection{Case Study}
 In Fig.~\ref{fig:case_study}, we deliver illustrative case examples from the test set of BMELD (En$\Rightarrow$Zh) to compare translations generated by different models.

\vspace{5pt}
\noindent\textbf{Dialogue Coherence.}
In the first example of Fig.~\ref{fig:case_study}, all contrast models translate the word ``\textit{game}'' into its surface meaning ``\textit{yóu xì}'' in Chinese. However, considering that the word ``\textit{antique}'' in dialogue history generally refers to physical assets rather than virtual objects, what the speaker $s1$ really means is ``\textit{yóu xì jī}'' (``\textit{arcade game machine}'') as in the reference, which is correctly translated by our ``Flat-NCT+MMT'' model. From the second example, we find that the translations generated by all contrast models neglect the crucial item ``\textit{boat}'' (``\textit{chuán}'') inside the dialogue. On the contrary, our model ``Flat-NCT+MMT'' successfully generates the translation of ``\textit{boat}'' that only exists in dialogue history context but not in the current utterance, which makes the whole translated utterance more coherent to the whole dialogue.

For the above two examples, the underlying reason for our model to generate more coherent translations is that the UD task in our proposed training framework introduces the modelling of dialogue coherence into the NCT model.

\vspace{5pt}
\noindent\textbf{Speaker Characteristic.}
We also observe that the translation generated by our model ``Flat-NCT+MMT'' can better preserve the characteristic of its original speaker. Specifically, in the second example of Fig.~\ref{fig:case_study}, the speaker $s1$ is highly excited and obviously in a tone of showing off. Consequently, our model converts the translation of the second ``\textit{What?}'' from its Chinese surface meaning ``\textit{shén me?}'' into a more speaker-consistent Chinese expression ``\textit{bù xìn?}'' (actually means ``\textit{don't you believe?}''), which makes the translated utterance more vivid and closer to the reference as well. This may be credited to the SD task that introduces the modelling of speaker characteristic into the NCT model during training.

The above case examples indicate that our proposed training framework makes the NCT model more capable of capturing important conversational properties of dialogue context, showing its superiority over other contrast models.

\section{Related Work}
\label{sec:related_work}
The most related work to ours include the studies of neural chat translation and context-aware NMT, which will be described in the following subsections.
\subsection{Neural Chat Translation}
Due to the lack of publicly available annotated bilingual dialogues, there are only few relevant studies on this task. To address the data scarcity issue, some researches \cite{DBLP:conf/lrec/WangZTWL16,DBLP:conf/wmt/MarufMH18,DBLP:conf/apsipa/ZhangZ19} design methods to automatically construct subtitles corpus that may contain low-quality bilingual dialogue utterances. Recently, Farajian et al., \cite{DBLP:conf/wmt/FarajianLMMH20} organize the competition of WMT20 shared task on chat translation and first provide a chat corpus post-edited by human annotators. In the competition, the submitted NCT systems \cite{DBLP:conf/wmt/BerardCNP20,DBLP:conf/wmt/MohammedAA20,DBLP:conf/wmt/WangTWDDS20} are trained with some typical engineering techniques such as ensemble for higher performances. All these systems adhere to the conventional two-stage pretrain-finetune paradigm, mainly including fine-tuning the existing models or using the large pre-trained language models such as BERT \cite{DBLP:conf/naacl/DevlinCLT19}. During pre-training on the large-scale parallel corpus, they either use all the available data or adopt data selection methods to select more in-domain data for training. More recently, Wang et al. \cite{wang2021autocorrect} propose to utilize context to translate dialogue utterances along with jointly identifying omission and typos in the process of translating. Different from these work, our proposed framework focuses on utilizing additional monolingual dialogues and introducing an intermediate stage to alleviate training discrepancy. 

\subsection{Context-aware NMT}
In a sense, NCT can be viewed as a special case of context-aware NMT that has recently attracted much attention \cite{DBLP:journals/corr/JeanLFC17,DBLP:conf/discomt/TiedemannS17,DBLP:conf/emnlp/WangTWL17,agrawal2018contextual,DBLP:conf/emnlp/ZhangLSZXZL18,DBLP:conf/ijcai/ZhengYHCB20,DBLP:conf/emnlp/KangZZZ20,DBLP:conf/acl/LiLWJXZLL20,DBLP:conf/acl/MaZZ20}. Typically, dominant approaches mainly resorted to extending extend conventional NMT models by incorporating cross-sentence global context, which can be roughly classified into two common categories: 1) concatenating the context and the current sentence to construct context-aware inputs \cite{agrawal2018contextual,DBLP:conf/discomt/TiedemannS17,DBLP:conf/acl/MaZZ20}; 2) using additional modules or modifying model architectures to encode context sentences \cite{DBLP:journals/corr/JeanLFC17,DBLP:conf/emnlp/WangTWL17,DBLP:conf/acl/VoitaSST18,DBLP:conf/emnlp/ZhangLSZXZL18,DBLP:conf/ijcai/ZhengYHCB20}. Besides, Kang et al. \cite{DBLP:conf/emnlp/KangZZZ20} considered the relevance of context sentences to the source sentence in document-level NMT and proposed to dynamically select relevant contextual sentences for each source sentence via reinforcement learning. Although these context-aware NMT models can be directly applied to the scenario of chat translation, they cannot overcome the previously-mentioned limitations of NCT models.

Apart from improving context-aware NMT models, some researches \cite{DBLP:conf/acl/VoitaST19a,DBLP:conf/acl/LiLWJXZLL20} investigated the effect of context in the process of translation. Voita et al., \cite{DBLP:conf/acl/VoitaST19a} concerned about the issue that the plausible translations of isolated sentences produced by context-agnostic NMT systems often end up being inconsistent with each other in a document. They investigated various linguistic phenomena and identified deixis, ellipsis and lexical cohesion as three main sources of inconsistency. Li et al. \cite{DBLP:conf/acl/LiLWJXZLL20} looked into how the contexts bring improvements to conventional document-level multi-encoder NMT models. They found that the context encoder behaves as a noise generator and improves NMT models with robust training especially when the training data is small. 

Not only are these findings suitable for context-aware NMT models in document translation, they also inspire follow-up researches on NCT to explore better ways of utilizing dialogue contexts such as explicitly modelling conversational properties of utterances.

\vspace{-5pt}
\section{Conclusion}
\label{sec:conclusion}
In this paper, we have proposed a multi-task multi-stage transitional training framework for neural chat translation, where an NCT model is trained using the bilingual chat translation dataset and additional monolingual dialogues. Particularly, we design UD and SD tasks to incorporate the modelling of dialogue coherence and speaker characteristic into the NCT model, respectively. Moreover, our proposed training framework consists of three stages: 1) sentence-level pre-training on large-scale parallel corpus; 2) intermediate training with auxiliary tasks using additional monolingual dialogues; 3) context-aware fine-tuning with with gradual transition. Experimental results and in-depth analysis demonstrate the effectiveness of our proposed training framework.



%



\ifCLASSOPTIONcompsoc
  \section*{Acknowledgments}
\else
  \section*{Acknowledgment}
\fi
The project was supported by National Natural Science Foundation of China (No. 62036004, No. 61672440), Natural Science Foundation of Fujian Province of China (No. 2020J06001), and Youth Innovation Fund of Xiamen (No. 3502Z20206059). We also thank the reviewers for their insightful comments. Work done while Chulun Zhou was an intern at Pattern Recognition Center, WeChat AI, Tencent Inc., Beijing, China.

\ifCLASSOPTIONcaptionsoff
  \newpage
\fi

\bibliography{mybib}

\begin{thebibliography}{10}
\providecommand{\url}[1]{#1}
\csname url@samestyle\endcsname
\providecommand{\newblock}{\relax}
\providecommand{\bibinfo}[2]{#2}
\providecommand{\BIBentrySTDinterwordspacing}{\spaceskip=0pt\relax}
\providecommand{\BIBentryALTinterwordstretchfactor}{4}
\providecommand{\BIBentryALTinterwordspacing}{\spaceskip=\fontdimen2\font plus
\BIBentryALTinterwordstretchfactor\fontdimen3\font minus
  \fontdimen4\font\relax}
\providecommand{\BIBforeignlanguage}[2]{{%
\expandafter\ifx\csname l@#1\endcsname\relax
\typeout{** WARNING: IEEEtran.bst: No hyphenation pattern has been}%
\typeout{** loaded for the language `#1'. Using the pattern for}%
\typeout{** the default language instead.}%
\else
\language=\csname l@#1\endcsname
\fi
#2}}
\providecommand{\BIBdecl}{\relax}
\BIBdecl

\bibitem{DBLP:conf/nips/SutskeverVL14}
I.~Sutskever, O.~Vinyals, and Q.~V. Le, ``Sequence to sequence learning with
  neural networks,'' in \emph{Advances in Neural Information Processing Systems
  27: Annual Conference on Neural Information Processing Systems}, 2014, pp.
  3104--3112.

\bibitem{DBLP:journals/corr/BahdanauCB14}
D.~Bahdanau, K.~Cho, and Y.~Bengio, ``Neural machine translation by jointly
  learning to align and translate,'' in \emph{3rd International Conference on
  Learning Representations}, 2015.

\bibitem{Vaswani:nips17}
A.~Vaswani, N.~Shazeer, N.~Parmar, J.~Uszkoreit, L.~Jones, A.~N. Gomez,
  L.~Kaiser, and I.~Polosukhin, ``Attention is all you need,'' in
  \emph{Advances in Neural Information Processing Systems 30: Annual Conference
  on Neural Information Processing Systems}, 2017, pp. 4831--4836.

\bibitem{DBLP:conf/discomt/TiedemannS17}
J.~Tiedemann and Y.~Scherrer, ``Neural machine translation with extended
  context,'' in \emph{Proceedings of the Third Workshop on Discourse in Machine
  Translation, DiscoMT@EMNLP}, 2017, pp. 82--92.

\bibitem{DBLP:conf/acl/HaffariM18}
S.~Maruf and G.~Haffari, ``Document context neural machine translation with
  memory networks,'' in \emph{Proceedings of the 56th Annual Meeting of the
  Association for ComputationalLinguistics}, 2018, pp. 1275--1284.

\bibitem{DBLP:conf/naacl/BawdenSBH18}
R.~Bawden, R.~Sennrich, A.~Birch, and B.~Haddow, ``Evaluating discourse
  phenomena in neural machine translation,'' in \emph{Proceedings of Conference
  of the North American Chapter of the Association for Computational
  Linguistics}, 2018, pp. 1304--1313.

\bibitem{DBLP:conf/emnlp/WerlenRPH18}
L.~M. Werlen, D.~Ram, N.~Pappas, and J.~Henderson, ``Document-level neural
  machine translation with hierarchical attention networks,'' in
  \emph{Proceedings of the Conference on Empirical Methods in Natural Language
  Processing}, 2018, pp. 2947--2954.

\bibitem{DBLP:journals/tacl/TuLSZ18}
Z.~Tu, Y.~Liu, S.~Shi, and T.~Zhang, ``Learning to remember translation history
  with a continuous cache,'' \emph{Trans. Assoc. Comput. Linguistics}, vol.~6,
  pp. 407--420, 2018.

\bibitem{DBLP:conf/acl/VoitaSST18}
E.~Voita, P.~Serdyukov, R.~Sennrich, and I.~Titov, ``Context-aware neural
  machine translation learns anaphora resolution,'' in \emph{Proceedings of the
  56th Annual Meeting of the Association for Computational Linguistics}, 2018,
  pp. 1264--1274.

\bibitem{DBLP:conf/acl/VoitaST19a}
E.~Voita, R.~Sennrich, and I.~Titov, ``When a good translation is wrong in
  context: Context-aware machine translation improves on deixis, ellipsis, and
  lexical cohesion,'' in \emph{Proceedings of the 57th Conference of the
  Association for Computational Linguistics}, 2019, pp. 1198--1212.

\bibitem{DBLP:conf/emnlp/VoitaST19b}
\BIBentryALTinterwordspacing
------, ``Context-aware monolingual repair for neural machine translation,'' in
  \emph{Proceedings of the Conference on Empirical Methods in Natural Language
  Processing and the 9th International Joint Conference on Natural Language
  Processing}, 2019, pp. 877--886. [Online]. Available:
  \url{https://doi.org/10.18653/v1/D19-1081}
\BIBentrySTDinterwordspacing

\bibitem{DBLP:conf/emnlp/WangTWS19}
L.~Wang, Z.~Tu, X.~Wang, and S.~Shi, ``One model to learn both: Zero pronoun
  prediction and translation,'' in \emph{Proceedings of the Conference on
  Empirical Methods in Natural Language Processing and the 9th International
  Joint Conference on Natural Language Processing}, 2019, pp. 921--930.

\bibitem{DBLP:conf/naacl/MarufMH19}
\BIBentryALTinterwordspacing
S.~Maruf, A.~F.~T. Martins, and G.~Haffari, ``Selective attention for
  context-aware neural machine translation,'' in \emph{Proceedings of
  Conference of the North American Chapter of the Association for Computational
  Linguistics}, J.~Burstein, C.~Doran, and T.~Solorio, Eds., 2019, pp.
  3092--3102. [Online]. Available: \url{https://doi.org/10.18653/v1/n19-1313}
\BIBentrySTDinterwordspacing

\bibitem{DBLP:conf/acl/MaZZ20}
S.~Ma, D.~Zhang, and M.~Zhou, ``A simple and effective unified encoder for
  document-level machine translation,'' in \emph{Proceedings of the 58th Annual
  Meeting of the Association for Computational Linguistics}, 2020, pp.
  3505--3511.

\bibitem{DBLP:conf/naacl/DevlinCLT19}
J.~Devlin, M.~Chang, K.~Lee, and K.~Toutanova, ``{BERT:} pre-training of deep
  bidirectional transformers for language understanding,'' in \emph{Proceedings
  of Conference of the North American Chapter of the Association for
  Computational Linguistics}, 2019, pp. 4171--4186.

\bibitem{DBLP:conf/coling/KuangXLZ18}
S.~Kuang, D.~Xiong, W.~Luo, and G.~Zhou, ``Modeling coherence for neural
  machine translation with dynamic and topic caches,'' in \emph{Proceedings of
  the 27th International Conference on Computational Linguistics}, 2018, pp.
  596--606.

\bibitem{DBLP:conf/emnlp/WangFWZ19}
W.~Wang, S.~Feng, D.~Wang, and Y.~Zhang, ``Answer-guided and semantic coherent
  question generation in open-domain conversation,'' in \emph{Proceedings of
  the 2019 Conference on Empirical Methods in Natural Language Processing and
  the 9th International Joint Conference on Natural Language Processing}, 2019,
  pp. 5065--5075.

\bibitem{DBLP:conf/aaai/XiongH0W19}
H.~Xiong, Z.~He, H.~Wu, and H.~Wang, ``Modeling coherence for discourse neural
  machine translation,'' in \emph{The Thirty-Third {AAAI} Conference on
  Artificial Intelligence, {AAAI} 2019, The Thirty-First Innovative
  Applications of Artificial Intelligence Conference, {IAAI} 2019, The Ninth
  {AAAI} Symposium on Educational Advances in Artificial Intelligence}, 2019,
  pp. 7338--7345.

\bibitem{DBLP:conf/ijcai/Wang019b}
T.~Wang and X.~Wan, ``{T-CVAE:} transformer-based conditioned variational
  autoencoder for story completion,'' in \emph{Proceedings of the Twenty-Eighth
  International Joint Conference on Artificial Intelligence}, S.~Kraus, Ed.,
  2019, pp. 5233--5239.

\bibitem{DBLP:conf/emnlp/HuangYQLL20}
L.~Huang, Z.~Ye, J.~Qin, L.~Lin, and X.~Liang, ``{GRADE:} automatic
  graph-enhanced coherence metric for evaluating open-domain dialogue
  systems,'' in \emph{Proceedings of the 2020 Conference on Empirical Methods
  in Natural Language Processing}, 2020, pp. 9230--9240.

\bibitem{DBLP:conf/wmt/WangTWDDS20}
L.~Wang, Z.~Tu, X.~Wang, L.~Ding, L.~Ding, and S.~Shi, ``Tencent {AI} lab
  machine translation systems for {WMT20} chat translation task,'' in
  \emph{Proceedings of the Fifth Conference on Machine Translation, WMT@EMNLP},
  2020, pp. 483--491.

\bibitem{DBLP:conf/acl/SennrichHB16a}
R.~Sennrich, B.~Haddow, and A.~Birch, ``Neural machine translation of rare
  words with subword units,'' in \emph{Proceedings of the 54th Annual Meeting
  of the Association for Computational Linguistics}, 2016.

\bibitem{DBLP:conf/emnlp/ByrneKSNGDYDKC19}
B.~Byrne, K.~Krishnamoorthi, C.~Sankar, A.~Neelakantan, B.~Goodrich,
  D.~Duckworth, S.~Yavuz, A.~Dubey, K.~Kim, and A.~Cedilnik, ``Taskmaster-1:
  Toward a realistic and diverse dialog dataset,'' in \emph{Proceedings of the
  2019 Conference on Empirical Methods in Natural Language Processing and the
  9th International Joint Conference on Natural Language Processing}, 2019, pp.
  4515--4524.

\bibitem{DBLP:conf/wmt/FarajianLMMH20}
M.~A. Farajian, A.~V. Lopes, A.~F.~T. Martins, S.~Maruf, and G.~Haffari,
  ``Findings of the {WMT} 2020 shared task on chat translation,'' in
  \emph{Proceedings of the Fifth Conference on Machine Translation, WMT@EMNLP},
  2020, pp. 65--75.

\bibitem{DBLP:conf/acl/PoriaHMNCM19}
S.~Poria, D.~Hazarika, N.~Majumder, G.~Naik, E.~Cambria, and R.~Mihalcea,
  ``{MELD:} {A} multimodal multi-party dataset for emotion recognition in
  conversations,'' in \emph{Proceedings of the 57th Conference of the
  Association for Computational Linguistics}, 2019, pp. 527--536.

\bibitem{DBLP:conf/wmt/MarufMH18}
S.~Maruf, A.~F.~T. Martins, and G.~Haffari, ``Contextual neural model for
  translating bilingual multi-speaker conversations,'' in \emph{Proceedings of
  the Third Conference on Machine Translation: Research Papers}, 2018, pp.
  101--112.

\bibitem{DBLP:conf/emnlp/ZhangLSZXZL18}
J.~Zhang, H.~Luan, M.~Sun, F.~Zhai, J.~Xu, M.~Zhang, and Y.~Liu, ``Improving
  the transformer translation model with document-level context,'' in
  \emph{Proceedings of the Conference on Empirical Methods in Natural Language
  Processing}, 2018, pp. 533--542.

\bibitem{DBLP:conf/amta/TanZHCWSLL20}
Z.~Tan, J.~Zhang, X.~Huang, G.~Chen, S.~Wang, M.~Sun, H.~Luan, and Y.~Liu,
  ``{THUMT:} an open-source toolkit for neural machine translation,'' in
  \emph{Proceedings of the 14th Conference of the Association for Machine
  Translation in the Americas}, 2020, pp. 116--122.

\bibitem{liang-etal-2021-modeling}
\BIBentryALTinterwordspacing
Y.~Liang, F.~Meng, Y.~Chen, J.~Xu, and J.~Zhou, ``Modeling bilingual
  conversational characteristics for neural chat translation,'' in
  \emph{Proceedings of ACL}, Aug. 2021, pp. 5711--5724. [Online]. Available:
  \url{https://aclanthology.org/2021.acl-long.444}
\BIBentrySTDinterwordspacing

\bibitem{DBLP:journals/corr/KingmaB14}
D.~P. Kingma and J.~Ba, ``Adam: {A} method for stochastic optimization,'' in
  \emph{3rd International Conference on Learning Representations}, 2015.

\bibitem{DBLP:conf/emnlp/Koehn04}
P.~Koehn, ``Statistical significance tests for machine translation
  evaluation,'' in \emph{Proceedings of the 2004 Conference on Empirical
  Methods in Natural Language Processing}, 2004, pp. 388--395.

\bibitem{liang-etal-2022-msctd}
\BIBentryALTinterwordspacing
Y.~Liang, F.~Meng, J.~Xu, Y.~Chen, and J.~Zhou, ``{MSCTD}: A multimodal
  sentiment chat translation dataset,'' in \emph{Proceedings of ACL}.\hskip 1em
  plus 0.5em minus 0.4em\relax Dublin, Ireland: Association for Computational
  Linguistics, May 2022, pp. 2601--2613. [Online]. Available:
  \url{https://aclanthology.org/2022.acl-long.186}
\BIBentrySTDinterwordspacing

\bibitem{DBLP:conf/ijcai/LapataB05}
M.~Lapata and R.~Barzilay, ``Automatic evaluation of text coherence: Models and
  representations,'' in \emph{Proceedings of the Nineteenth International Joint
  Conference on Artificial Intelligence}, 2005, pp. 1085--1090.

\bibitem{DBLP:journals/corr/abs-1301-3781}
T.~Mikolov, K.~Chen, G.~Corrado, and J.~Dean, ``Efficient estimation of word
  representations in vector space,'' in \emph{1st International Conference on
  Learning Representations}, 2013.

\bibitem{DBLP:conf/wmt/BaoSSLC20}
C.~Bao, Y.~Shiue, C.~Song, J.~Li, and M.~Carpuat, ``The university of
  maryland's submissions to the wmt20 chat translation task: Searching for more
  data to adapt discourse-aware neural machine translation,'' in
  \emph{Proceedings of the Fifth Conference on Machine Translation, WMT@EMNLP},
  2020, pp. 456--461.

\bibitem{doi:10.1177/001316447303300309}
\BIBentryALTinterwordspacing
J.~L. Fleiss and J.~Cohen, ``The equivalence of weighted kappa and the
  intraclass correlation coefficient as measures of reliability,''
  \emph{Educational and Psychological Measurement}, pp. 613--619, 1973.
  [Online]. Available: \url{https://doi.org/10.1177/001316447303300309}
\BIBentrySTDinterwordspacing

\bibitem{DBLP:conf/lrec/WangZTWL16}
L.~Wang, X.~Zhang, Z.~Tu, A.~Way, and Q.~Liu, ``Automatic construction of
  discourse corpora for dialogue translation,'' in \emph{Proceedings of the
  Tenth International Conference on Language Resources and Evaluation}, 2016.

\bibitem{DBLP:conf/apsipa/ZhangZ19}
L.~Zhang and Q.~Zhou, ``Automatically annotate {TV} series subtitles for
  dialogue corpus construction,'' in \emph{2019 Asia-Pacific Signal and
  Information Processing Association Annual Summit and Conference}, 2019, pp.
  1029--1035.

\bibitem{DBLP:conf/wmt/BerardCNP20}
A.~Berard, I.~Calapodescu, V.~Nikoulina, and J.~Philip, ``Naver labs europe's
  participation in the robustness, chat, and biomedical tasks at {WMT} 2020,''
  in \emph{Proceedings of the Fifth Conference on Machine Translation,
  WMT@EMNLP}, 2020, pp. 462--472.

\bibitem{DBLP:conf/wmt/MohammedAA20}
R.~Mohammed, M.~Al{-}Ayyoub, and M.~Abdullah, ``{JUST} system for {WMT20} chat
  translation task,'' in \emph{Proceedings of the Fifth Conference on Machine
  Translation, WMT@EMNLP}, 2020, pp. 479--482.

\bibitem{wang2021autocorrect}
T.~Wang, C.~Zhao, M.~Wang, L.~Li, and D.~Xiong, ``Autocorrect in the process of
  translation -- multi-task learning improves dialogue machine translation,''
  2021.

\bibitem{DBLP:journals/corr/JeanLFC17}
S.~Jean, S.~Lauly, O.~Firat, and K.~Cho, ``Does neural machine translation
  benefit from larger context?'' \emph{CoRR}, 2017.

\bibitem{DBLP:conf/emnlp/WangTWL17}
L.~Wang, Z.~Tu, A.~Way, and Q.~Liu, ``Exploiting cross-sentence context for
  neural machine translation,'' in \emph{Proceedings of the Conference on
  Empirical Methods in Natural Language Processing}, 2017, pp. 2826--2831.

\bibitem{agrawal2018contextual}
R.~R. Agrawal, M.~Turchi, and M.~Negri, ``Contextual handling in neural machine
  translation: Look behind, ahead and on both sides,'' in \emph{21st Annual
  Conference of the European Association for Machine Translation}, 2018, pp.
  11--20.

\bibitem{DBLP:conf/ijcai/ZhengYHCB20}
Z.~Zheng, X.~Yue, S.~Huang, J.~Chen, and A.~Birch, ``Towards making the most of
  context in neural machine translation,'' in \emph{Proceedings of the
  Twenty-Ninth International Joint Conference on Artificial Intelligence},
  2020, pp. 3983--3989.

\bibitem{DBLP:conf/emnlp/KangZZZ20}
X.~Kang, Y.~Zhao, J.~Zhang, and C.~Zong, ``Dynamic context selection for
  document-level neural machine translation via reinforcement learning,'' in
  \emph{Proceedings of the 2020 Conference on Empirical Methods in Natural
  Language Processing}, 2020, pp. 2242--2254.

\bibitem{DBLP:conf/acl/LiLWJXZLL20}
B.~Li, H.~Liu, Z.~Wang, Y.~Jiang, T.~Xiao, J.~Zhu, T.~Liu, and C.~Li, ``Does
  multi-encoder help? {A} case study on context-aware neural machine
  translation,'' in \emph{Proceedings of the 58th Annual Meeting of the
  Association for Computational Linguistics}, 2020, pp. 3512--3518.

\end{thebibliography}
\bibliographystyle{IEEEtran}

%
\begin{IEEEbiography}[{\includegraphics[width=1in,height=1.25in,clip,keepaspectratio]{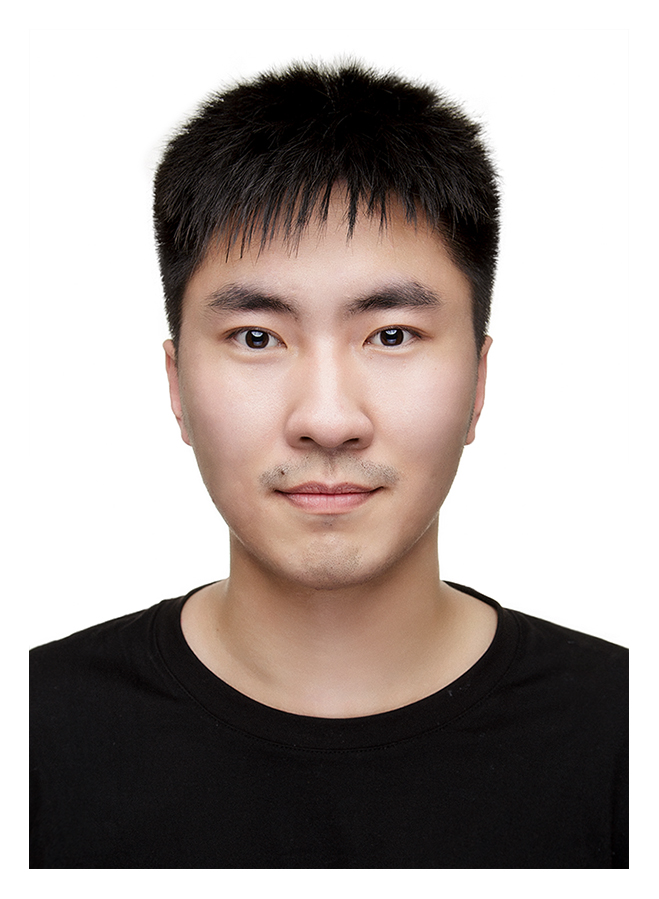}}]{Chulun Zhou}
Chulun Zhou received the M.S. degree from Xiamen University, Xiamen, China, in 2022. He is currently a researcher in Pattern Recognition Center, WeChat AI, Tencent Inc. His research interests include natural language processing, text generation and neural machine translation.
\end{IEEEbiography}

\begin{IEEEbiography}[{\includegraphics[width=1in,height=1.25in,clip,keepaspectratio]{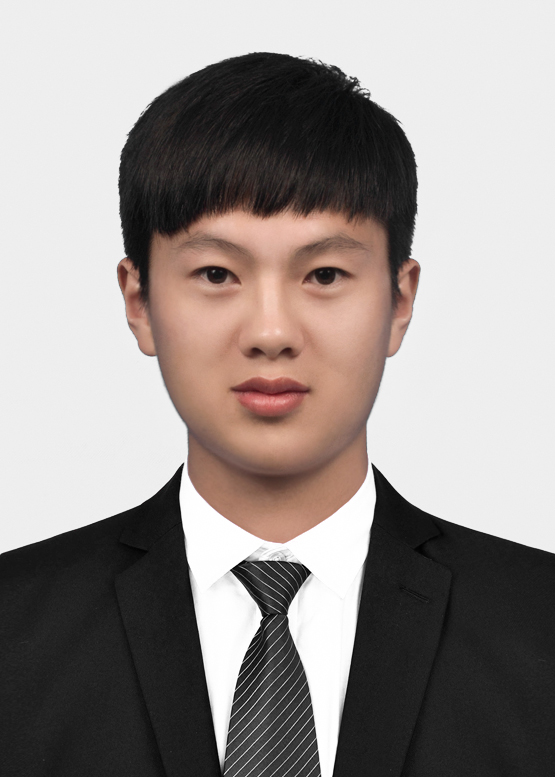}}]{Yunlong Liang}
Yunlong Liang received the B.S. degree from Hebei University of Technology, Tianjin, China, in 2018. He is currently a Ph.D. candidate in Beijing Jiaotong University, Beijing, China. His research interests include natural language processing, fine-grained sentiment analysis, emotional response generation, and machine translation.
\end{IEEEbiography}

\begin{IEEEbiography}[{\includegraphics[width=1in,height=1.25in,clip,keepaspectratio]{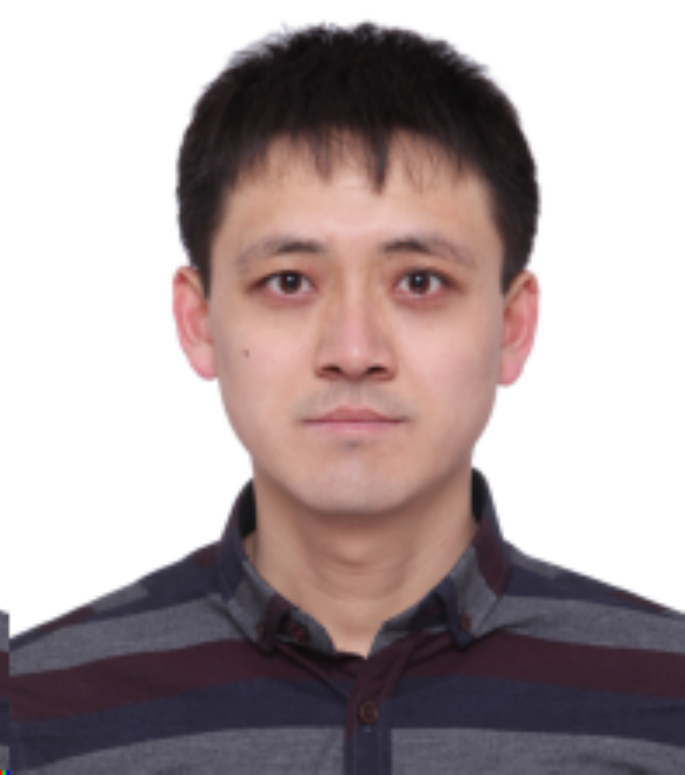}}]{Fandong Meng}
Fandong Meng received the Ph.D. degree in Chinese Academy of Sciences, and is now a principal researcher in Pattern Recognition Center, WeChat AI, Tencent Inc. His research interests include natural language processing, machine translation and dialogue system.
\end{IEEEbiography}
\vspace{-10pt}
\begin{IEEEbiography}[{\includegraphics[width=1in,height=1.25in,clip,keepaspectratio]{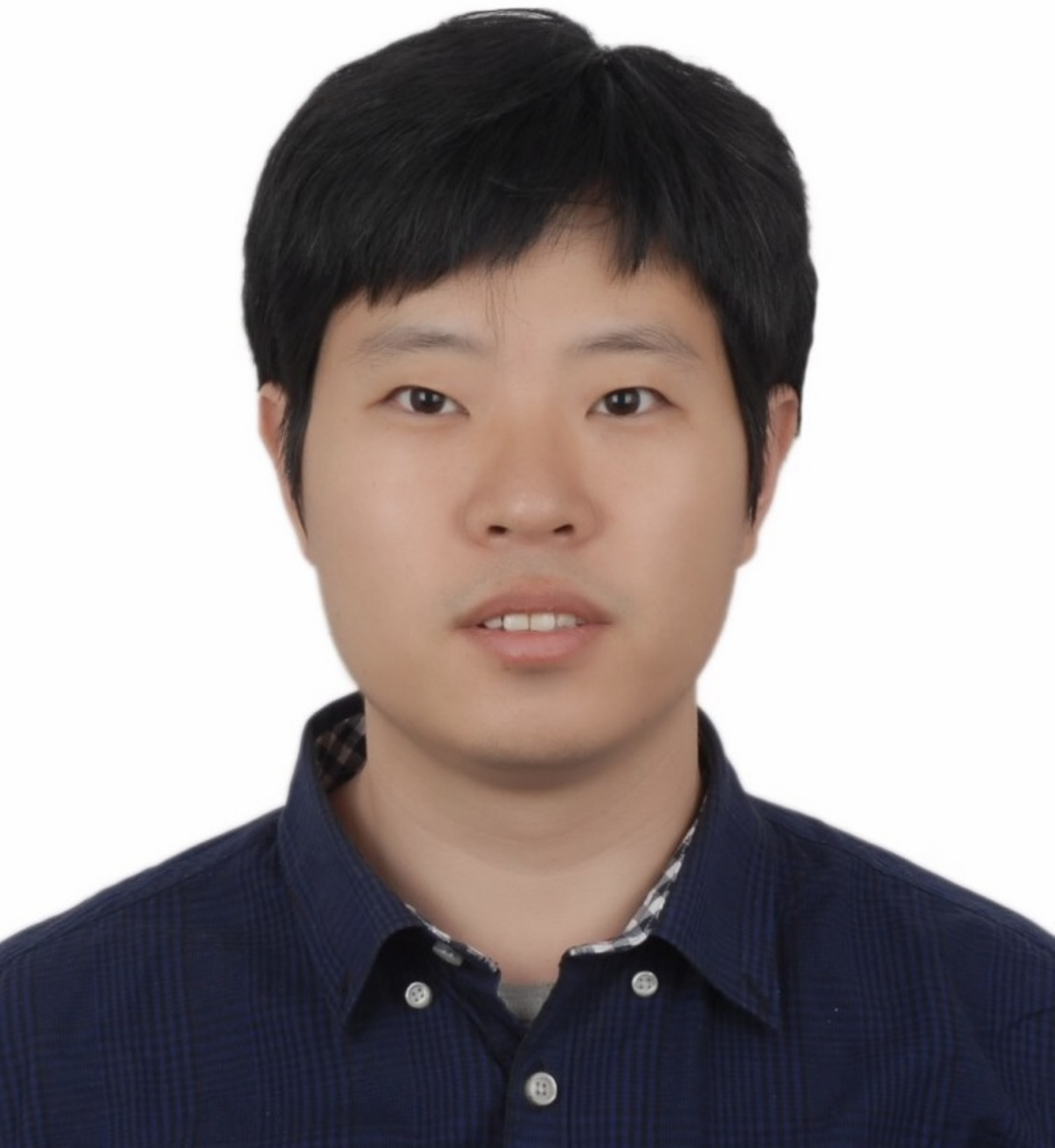}}]{Jie Zhou}
Jie Zhou received his bachelor degree from USTC in 2004 and his Ph.D. degree from Chinese Academy of Sciences in 2009, and is now a senior director of Pattern Recognition Center, WeChat AI, Tencent Inc. His research interests include natural language processing and machine learning.
\end{IEEEbiography}
\vspace{-10pt}
\begin{IEEEbiography}[{\includegraphics[width=1in,height=1.25in,clip,keepaspectratio]{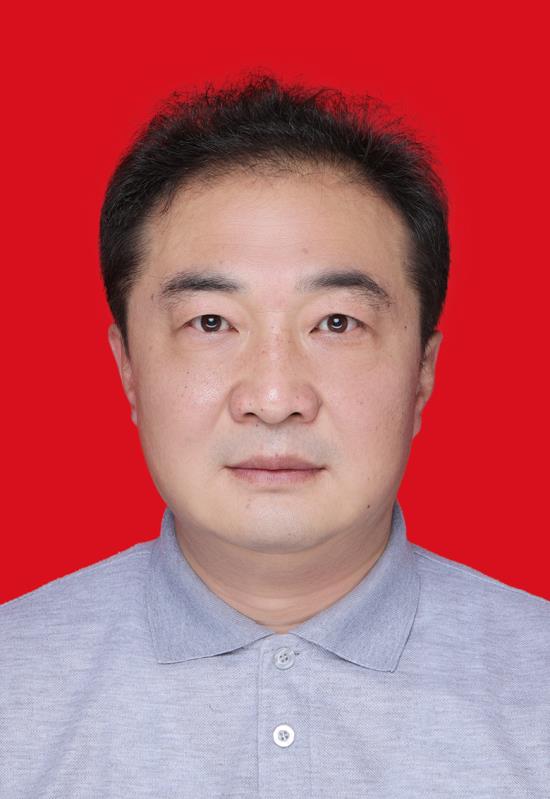}}]{Jinan Xu}
Jinan Xu received his PH.D. from Hokkaidao University in Japan in March 2006, and then he worked for NEC Lab. From August 2009 to now, he works for Beijing Jiaotong University as a professor, his main research fields include NLP, MT, Knowledge Graph, big data processing, etc. He is a senior member of CCF, and a member of CCF NLP Committee and Machine Translation Committee of Chinese Information Processing Society of China.
\end{IEEEbiography}
\vspace{-10pt}

\begin{IEEEbiography}[{\includegraphics[width=1in,height=1.25in,clip,keepaspectratio]{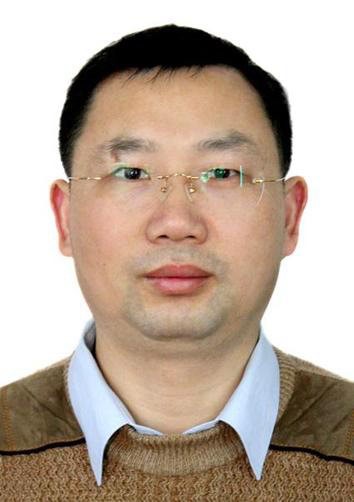}}]{Hongji Wang}
Hongji Wang received the Ph.D. degree at the Institute of Software, Chinese Academy of Sciences, and is now an associate professor at Xiamen University. His research interests include information security, software engineering, and intelligence analysis.
\end{IEEEbiography}
\vspace{-10pt}
\begin{IEEEbiography}[{\includegraphics[width=1in,height=1.25in,clip,keepaspectratio]{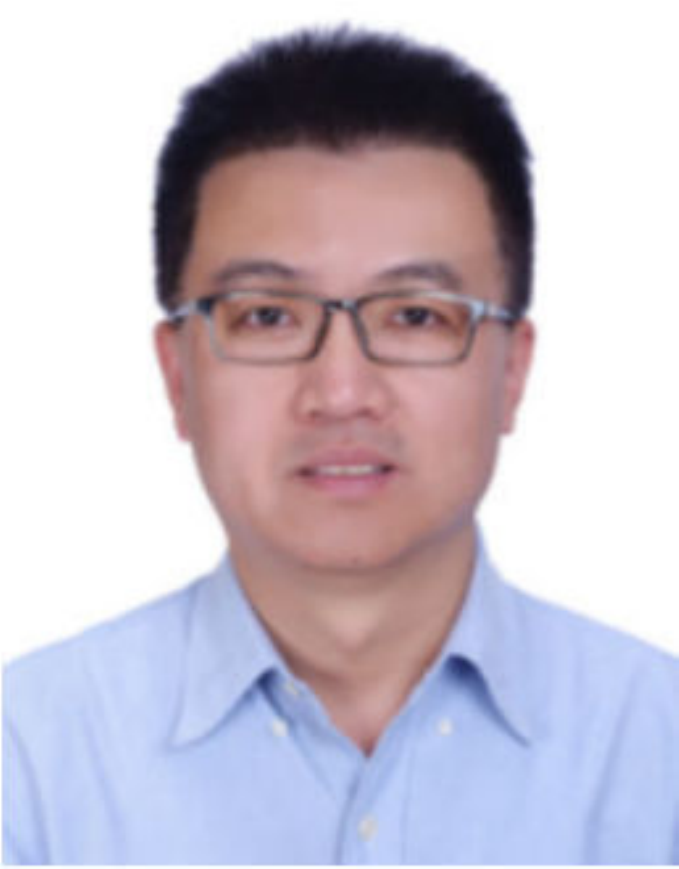}}]{Min Zhang}
Min Zhang (Member, IEEE) received the bachelors and Ph.D. degrees in computer science from the Harbin Institute of Technology, Harbin, China, in 1991 and 1997, respectively. He is currently a Distinguished Professor with the School of Computer Science and Technology, Soochow University, Suzhou, China. His current research interests include machine translation, natural language processing, and artificial intelligence.
\end{IEEEbiography}
\vspace{-10pt}

\begin{IEEEbiography}[{\includegraphics[width=1in,height=1.25in,clip,keepaspectratio]{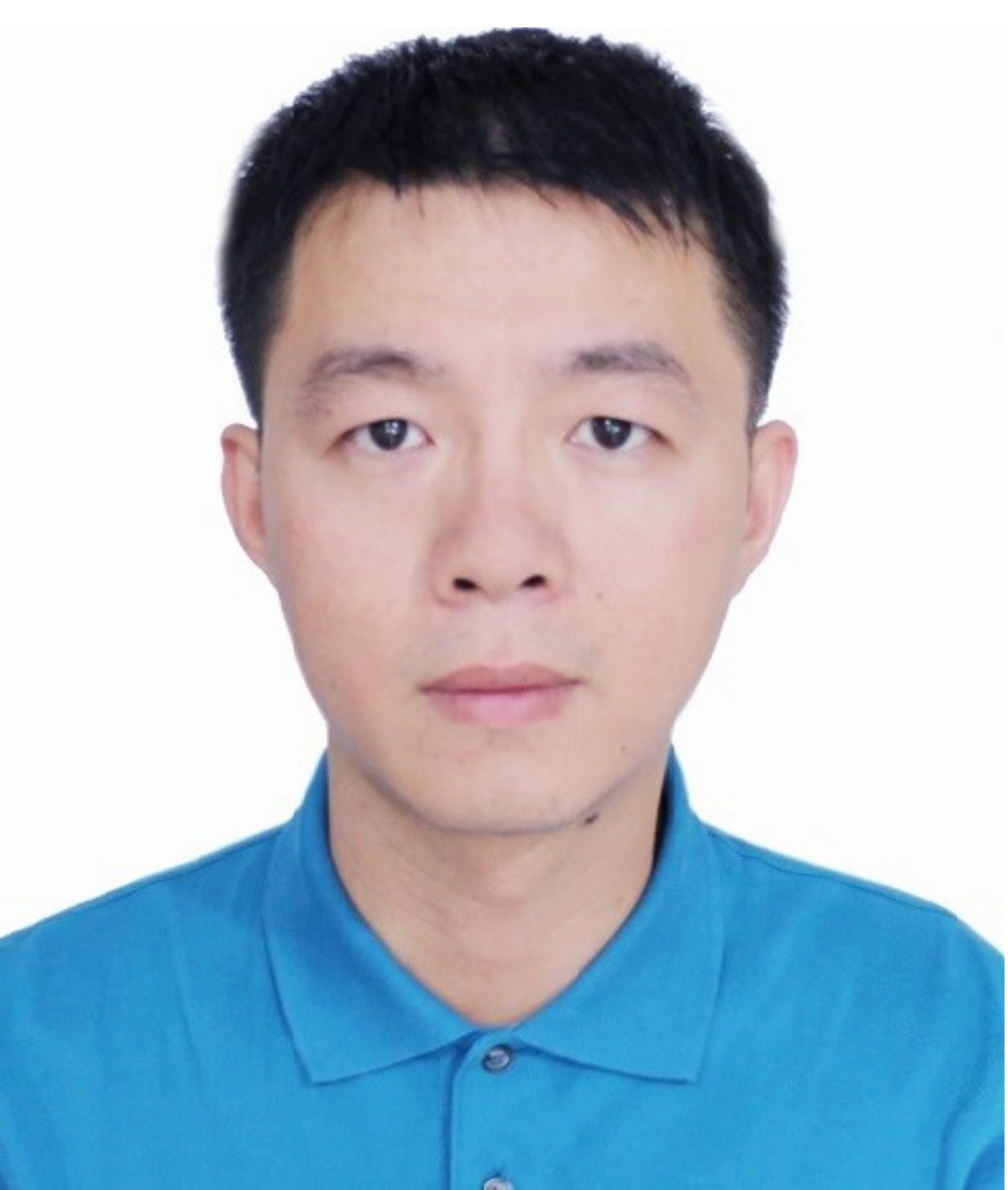}}]{Jinsong Su}
Jinsong Su was born in 1982. He received the Ph.D. degree in Chinese Academy of Sciences, and is now a professor in Xiamen University. His research interests include natural language processing, neural machine translation and text generation. He has served as the Area Co-Chair of the ACL 2021/2022, EMNLP 2019/2020/2022, COLING 2022, NLPCC 2018/2020.
\end{IEEEbiography}
\end{document}